%% file: main.tex
\pdfoutput=1
\documentclass[11pt]{article}

\usepackage[preprint]{acl}
\usepackage{times}
\usepackage{latexsym}
\usepackage[T1]{fontenc}
\usepackage[utf8]{inputenc}
\usepackage{microtype}
\usepackage{inconsolata}
\usepackage{graphicx}
\usepackage{enumitem}
\usepackage{bm}
\usepackage[textsize=tiny]{todonotes}
\usepackage{amsmath}
\usepackage{amssymb}
\usepackage{multirow}
\usepackage{booktabs}
\usepackage{colortbl}
\usepackage{textcomp}
\usepackage{makecell}
\usepackage{cleveref}
\makeatletter
\newcommand*\rel@kern[1]{\kern#1\dimexpr\macc@kerna}
\newcommand*\widebar[1]{%
  \begingroup
  \def\mathaccent##1##2{%
    \rel@kern{0.8}%
    \overline{\rel@kern{-0.8}\macc@nucleus\rel@kern{0.2}}%
    \rel@kern{-0.2}%
  }%
  \macc@depth\@ne
  \let\math@bgroup\@empty \let\math@egroup\macc@set@skewchar
  \mathsurround\z@ \frozen@everymath{\mathgroup\macc@group\relax}%
  \macc@set@skewchar\relax
  \let\mathaccentV\macc@nested@a
  \macc@nested@a\relax111{#1}%
  \endgroup
}
\makeatother

\title{Length Generalization of Causal Transformers without Position Encoding}

\author{
Jie Wang\textsuperscript{\rm 1}\thanks{\ \ Equal contribution.},
Tao Ji\textsuperscript{\rm 2}\footnotemark[1],
Yuanbin Wu\textsuperscript{\rm 1},\\
\textbf{
Hang Yan\textsuperscript{\rm 5},
Tao Gui\textsuperscript{\rm 3},
Qi Zhang\textsuperscript{\rm 2},
Xuanjing Huang\textsuperscript{\rm 2,4},
Xiaoling Wang\textsuperscript{\rm 1}}\\
  {$^1$ School of Computer Science, East China Normal University, Shanghai, China} \\
  {$^2$ School of Computer Science, Fudan University, Shanghai, China} \\
  {$^3$ Institute of Modern Languages and Linguistics, Fudan University, Shanghai, China} \\
  {$^4$ International Human Phenome Institutes, Shanghai, China}\ \ \ 
  {$^5$ Shanghai AI Lab} \\
  \texttt{jiewang.cs@stu.ecnu.edu.cn,
  taoji@fudan.edu.cn,
  ybwu@cs.ecnu.edu.cn
  }
}

\begin{document}
\maketitle
\begin{abstract}
  \input{sec/1_abstract}
\end{abstract}

\input{sec/2_intro}

\input{sec/3_method}

\input{sec/4_experiment}

\input{sec/5_related_work}

\input{sec/6_conclusion}

% \begin{quote}
% \begin{verbatim}
% \bibliography{anthology,custom}
% \end{verbatim}
% \end{quote}

\bibliography{anthology,custom}

\appendix
\input{sec/8_appendix}

\end{document}

%% file: sec/1_abstract.tex
Generalizing to longer sentences
is important for recent Transformer-based language models.
Besides algorithms manipulating explicit position features,
the success of 
Transformers without position encodings (NoPE)
provides a new way to overcome the challenge.
In this paper, we study the length generalization property 
of NoPE.
We find that although NoPE can extend to longer sequences 
than the commonly used explicit position encodings, 
it still has a limited context length.
We identify a connection between the failure of NoPE’s 
generalization and the distraction of attention distributions.
We propose a parameter-efficient tuning for 
searching attention heads' best temperature hyper-parameters, 
which substantially expands NoPE's context size.
Experiments on long sequence language modeling, the synthetic
passkey retrieval task and real-world long context tasks show 
that NoPE can achieve competitive performances with state-of-the-art 
length generalization algorithms.
The source code is publicly 
accessible\footnote{$\ \ $\url{https://github.com/AntNLP/nope_head_scale}}.

% We study the problem of 
% We discover a connection between 
% We propose a new
% The results

%% file: sec/2_intro.tex
\section{Introduction}
\label{sec:intro}
%Causal Transformer without position encoding (NoPE).

Causal Transformer has been widely applied 
in modern language models. 
To help models recognize the correct ordering of words, 
it is common to configure Transformers
with \emph{explicit} position encodings
(e.g., the sinusoidal embeddings in the original
development of Transformer \cite{NIPS2017_3f5ee243}, 
the relative position encoding in T5 \cite{JMLR:v21:20-074},
and the rotary position encoding in GPT series \cite{DBLP:journals/corr/abs-2104-09864}).
The setup of position features provides
flexibility to include prior knowledge structure on 
describing distance,
but it also brings the problem of \emph{length generalization}:
language models trained with in-domain position features
can not handle longer sentences 
(i.e., those with out-of-domain position features)
in testing time.
%Extending the context window 
Generalizing to unseen sentence length
is crucial in many language model applications
like retrieval augmented language models \cite{DBLP:journals/jmlr/IzacardLLHPSDJRG23}, 
personalized language models \cite{wang2023rolellm},
language-model-based agents \cite{10.1145/3586183.3606763}.

% Nope, length generalization of Nope, challenges

Departing from the standard ways of encoding positions,
one may ask (following the principle of parsimony) that 
are the explicit position features necessary?
The answer is no.
Both empirically \cite{haviv-etal-2022-transformer} and theoretically \cite{chi-etal-2023-latent,kazemnejad2023the},
the casually masked Transformers are shown to be able to
successfully model languages
without any prior position encoding (\textbf{NoPE}).
The finding calls for a deeper understanding of
\emph{implicit} position information in 
Transformer-based language models,
and also inspires a new direction for length generalization:
\emph{without explicit position features, can NoPE generalize?}

\input{fig/fig_vis_entro}

\input{fig/fig_vis_uni_scale_entro}

In this paper, we study the length generalization property of NoPE.
Our main findings are,

\begin{itemize}[leftmargin=*]
    \item When extending to unseen sentence length, 
    NoPE has less performance loss.
    However, beyond a certain range, NoPE also fails to extend, 
    with no substantial difference observed when compared to explicit position encodings.
    %can be extended to longer sentences
    For example, NoPE can effectively extend the training length by 
    $20\%$ (from $2$K to $2.4$K, Figure \ref{fig:vis_entro}) 
    without a significant increase in perplexity.
    In contrast, the rotary position encoding (RoPE) is only capable of extending by $10\%$. 
    
    \item 
    We analyze the failure cases of NoPE's generalization
    and find that they always co-occur with 
    the distraction of attention distributions:
    the attention heads begin to allocate their weights to tokens evenly when NoPE's extension performance begins to collapse. 
    The connection between NoPE's generalization 
    and concentration of attention heads suggests controlling the behaviors of attention heads during length extension.
    % TODO XXX
    % possible theoretical results？We source this phenomenon 
    \item We show that by simply searching one temperature hyper-parameter, NoPE's length generalization 
    can be significantly improved. For example, by scaling the attention score by a factor of $1.2$, NoPE can immediately generalize to over $4$K tokens (Figure \ref{fig:vis_entro}).
    \item Moreover, we developed an advanced version of this 
    strategy by searching temperature parameters for each head, in the light that different layers and heads exhibit varied behaviors. 
    The procedure resembles a parameter-efficient fine-tuning,
    with an extremely small number of tunable parameters 
    ($704$ delta parameters over $1$B model parameters).
    We show that the proposed method can
    help NoPE to generalize further (Figure \ref{fig:vis_head_vs_uni}). %(results XXX).
\end{itemize}

% A similar skill has been applied in 
% Transformers with rotary position encoding \cite{chiang2022overcoming}\todo{ref YaRN?},
% but we show that it only helps get a marginal extension of length there. %(results XXX)

We conduct length generalization experiments on 
long sequence language modeling, 
synthetic tasks (passkey retrieval), and 
LongBench.
The results show that NoPE 
enjoys a competitive
extension performances to state-of-the-art 
length generalization methods for explicit
position encodings (e.g., 
PI \cite{chen2023extending}, 
YaRN \cite{peng2024yarn}).

% Our observations and methods
% - observations: the connection between failed generalization of Nope and difficulty of concentrating attention.
% - methods: head scaling (which could be recognized as an parameter efficient tuning)

% Experiments
% - PPL 
% - passkey retrieval

%% file: fig/fig_vis_entro.tex
\begin{figure}[t]
  \centering
  \includegraphics[width=\linewidth]{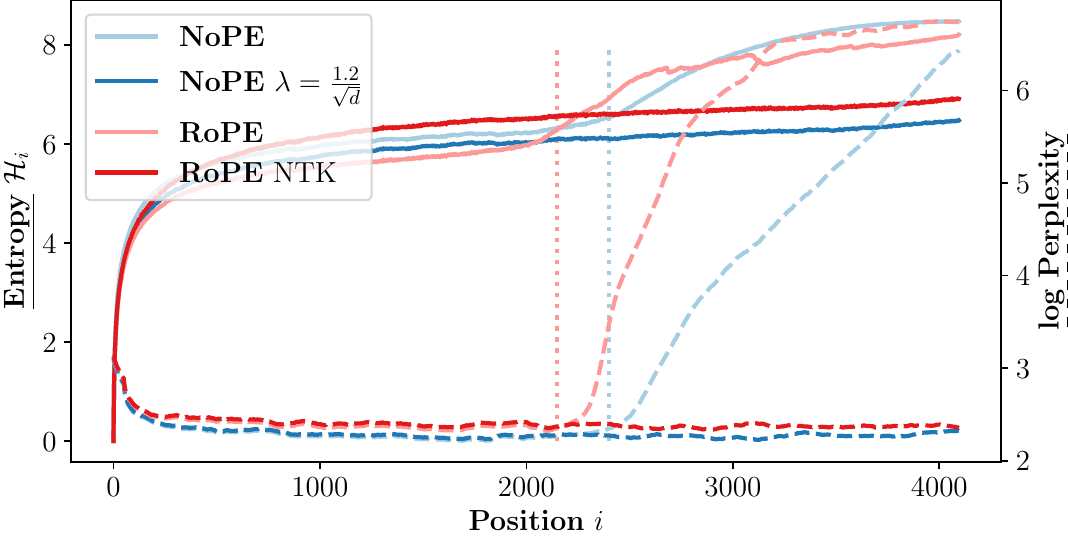}
  \caption{Length generalization from $2$K to $4$K. 
  For different testing lengths (or, positions of sequences), 
  dashed lines draw the log-perplexity of models
  (measured on validation set of the pre-training dataset),
  and solid lines represent the entropy of attention heads
  (averaged on all heads).}
  %\caption{TODO: Visualization of average entropy. (a) NoPE (b) NoPE+const scale (c) RoPE (d) RoPE+NTK or RoPE+YaRN}
  % \vspace{-0.5cm}
  \label{fig:vis_entro}
\end{figure}

%% file: fig/fig_vis_uni_scale_entro.tex
\begin{figure*}[t]
  \centering
  \includegraphics[width=0.49\linewidth]{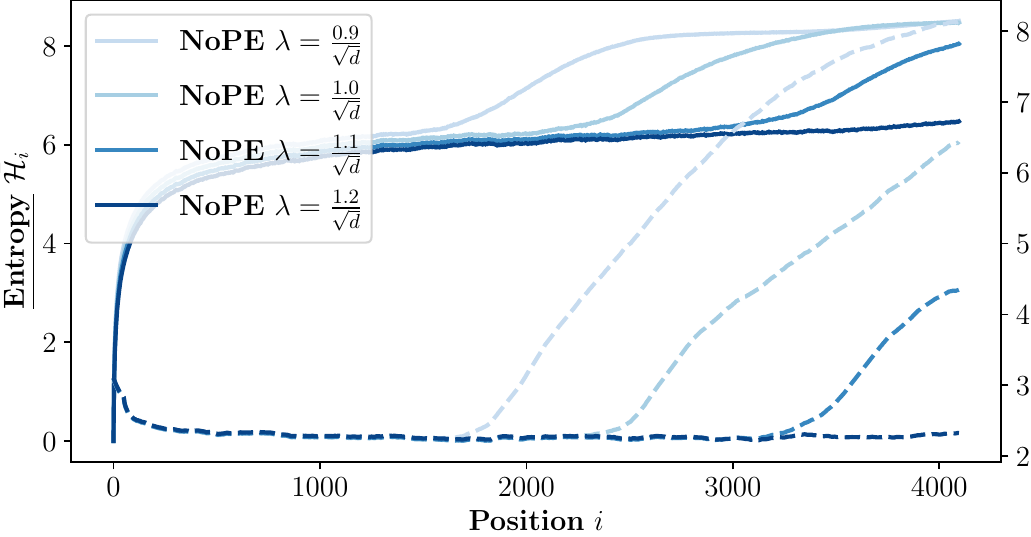}
  \includegraphics[width=0.49\linewidth]{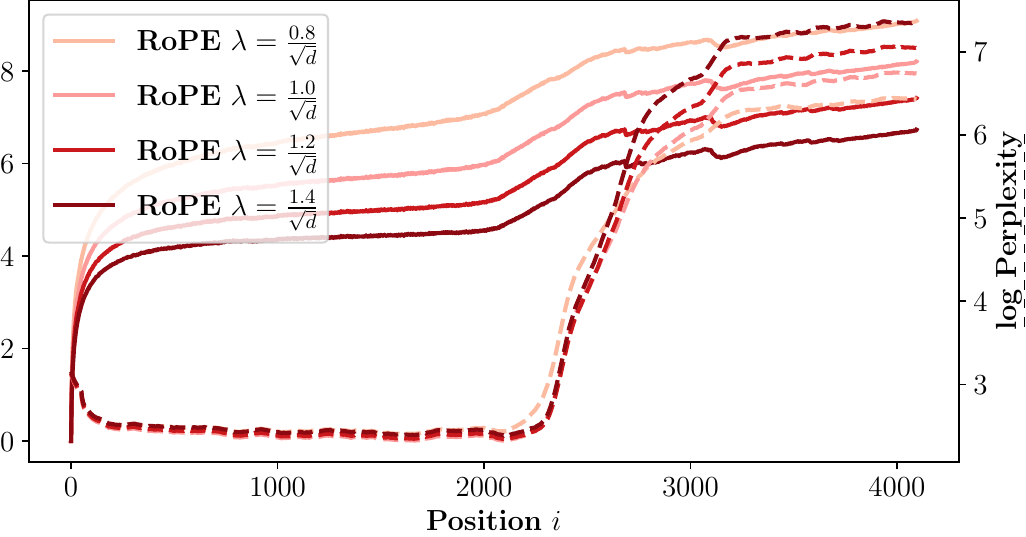}
  \caption{\textbf{UniformScale} modifies the temperature hyper-parameter of the $\mathrm{SoftMax}$ operator in self-attention layers (Left, NoPE; Right, RoPE). NoPE can generalize to longer context by merely scaling the softmax scores. However, this exact technique does not directly apply to RoPE models.}
  % \vspace{-0.5cm}
  \label{fig:vis_uni_entro}
\end{figure*}

%% file: sec/3_method.tex
\section{Length Generalization of NoPE}
\label{sec:method}

\subsection{Language Modeling with NoPE}
\label{sec:lm}

% Before diving into the length generalization problem, 
% we first briefly describe the NoPE models used in this paper.
% \footnote{For simplicity, 
% we refer NoPE to both the implicit way of encoding positions 
% and the language model trained without position encoding.}
% Our default NoPE is trained from the 
% TinyLlama \cite{zhang2024tinyllama} code base
% \footnote{\url{https://github.com/jzhang38/TinyLlama}},
% with training sequence length $L=2048$, $22$ layers of Transformer
% blocks, $32$ attention heads per-layer, $2048$ embedding size,
% and overall $1.1$B parameters. 
% The model is trained on Slimpajama \cite{cerebras2023slimpajama} joint with Starcoderdata \cite{li2023starcoder}  by $50$K steps ($\approx 100$B tokens).

Before diving into the length generalization problem, 
we first briefly describe the NoPE models used in this paper.
\footnote{For simplicity, 
we refer NoPE to both the implicit way of encoding positions 
and the language model trained without position encoding.}
Our default NoPE has $1.1$B parameters. It is trained from 
the TinyLlama \cite{zhang2024tinyllama} code base
\footnote{\url{https://github.com/jzhang38/TinyLlama}}, 
with training sequence length $L=2048$ and $50$K steps 
($\approx 100$B tokens). More details can be found in 
Section~\ref{ssec:nope}.

We also include the original TinyLlama model
which uses rotary position encoding (RoPE) for comparison.
By default, both models are trained with identical settings.

%The NoPE LMs in this paper are obtained from the open-source repository TinyLLama-RoPE and trained from scratch by removing PE, or initialized and then fine-tuned with the non-PE parameters from the pre-trained RoPE checkpoint.

\subsection{Length Generalization}
\label{sec:lg}

Given a language model (LM) with pre-trained maximal sequence length $L$, the goal of length generalization is to expand it to length $L^\prime > L$.
%the extension ratio $s \triangleq L^\prime/L$.
Length generalization can be tested in a zero-shot manner 
(``train short, test long'') or with some fine-tuning.

Figure \ref{fig:vis_entro} depicts language modeling 
performances of NoPE (and RoPE).
We can observe that, 
within the pre-training length ($L = 2048$),
NoPE has a similar performance as RoPE,
which agrees with existing works:
casual masking can implicitly encode the positions of a sequence
\citep{haviv-etal-2022-transformer,chi-etal-2023-latent}.

When the testing sequence length exceeds the training length,
we see that 
1) NoPE's length generalization error 
(light blue dashed line, measured with log-perplexity) 
is lower than RoPE (light red dashed line).
2) vanilla NoPE still has an increased perplexity 
than in-domain tests.
Therefore, though it is not a perfect solution, 
removing explicit position encoding can 
effectively reduce the length generalization error.
Next, we will try to find the reason 
for the failure of NoPE's length generalization,
and also develop algorithms for improving it.
%To further improve NoPE's 
%which call for a detailed analyses of fa

%but still suffer from weak length generalization outside the pre-training length ($>\!L$) \citep{kazemnejad2023the}. 

%Our goal is twofold, 
%\begin{enumerate}[noitemsep,leftmargin=24pt,topsep=0pt]
%    \item to analyze the reason for the failure of length generalization after excluding the influence of explicit PE, 
%    \item to explore length generalization methods for NoPE.
%\end{enumerate}

% \subsection{Visualization Tool: Attention Entropy}
%\subsection{Visualization: Attention Entropy}
%\subsection{Generalization and Attention}
\subsection{Extension? Attention!}
\label{ssec:vis_entro}

To analyze NoPE's generalization failure,
we first see that since
explicit position encodings have been dropped, the casual Transformer block is only left with three core modules,
the embedding layer, feed-forward layers, and self-attention layers.
The outputs of the former two modules are independent of their inputs' position in sequence (i.e., no matter which position, they always have the same output).
Therefore, multi-head attention layers become our main target.
%a position-dependent multi-head attention module and two position-independent modules, the word embedding module and the feed-forward module.
%The multi-head attention module is now becoming a logical primary focus.

We visualize the attention pattern of NoPE at different lengths.
%Specifically, we choose entropy to represent the attention pattern.
%Specifically, at each position $i$, we compute
Specifically, 
given a validation set with a size $n$ and
a target position $i$, we define the average attention entropy 
$\widebar{\mathcal{H}}_i$ at position $i$, as
%Here, $\widebar{\mathcal{H}}_i$ is the average of the entropy values $\mathcal{H}_i^{(h)}(x)$ across all samples for each attention head $h\in\{1, 2, \dots, m\}$.
\begin{align}
\widebar{\mathcal{H}}_i &= \frac{1}{n\!\times\!m} \sum_{x,h} \mathcal{H}_i^{(h)}(x) \\
\mathcal{H}_i^{(h)}(x) &= -\sum_{j=1}^{i} \alpha_{ij}^{(h)}(x) \cdot \log \alpha_{ij}^{(h)}(x)
% \alpha_{ij}^{(h)} &= \text{Softmax}_j(\bm{q}_i^{(h)} \cdot \bm{K}^{(h)})
\label{eq:entropy}
\end{align}
where $x$ is a sample, 
$\alpha_{ij}^{(h)}(x)$ is the attention probability of 
token $i$ focusing on token $j$ in the $h$-th attention head
($h\in\{1,2,...,m\}$),
$\mathcal{H}_i^{(h)}(x)$ is the entropy of the attention distribution $\alpha_{ij}^{(h)}(x)$ evaluated at position $i$.

% With $\bm{q}_i^{(h)}$ as the query vector and $\bm{K}^{(h)}$ as the matrix of all key vectors, the Softmax is applied over attention scores at position $j$ to obtain $\alpha_{ij}^{(h)}$.

%In this section, we introduce a visualization tool named attention entropy, which is an effective tool for measuring attention patterns.
%When the entropy is close to the lower bound (i.e., 0), it indicates more focused attention, and when it's close to the upper bound (i.e., $\log(i)$), it indicates more distracted attention.
%Where $i$ is the current generating position and also the number of tokens involved in the attention calculation.

The light solid lines in Figure~\ref{fig:vis_entro} show
the average entropy for NoPE (light blue) and RoPE (light red).
%with or without the length generalization method, respectively.
We can observe that,
%\textbf{Observation 1: for original NoPE and RoPE, successful generalization shows smooth $\widebar{\mathcal{H}}_i$, while failure in generalization shows a steep increase $\widebar{\mathcal{H}}_i$, i.e., distracted attention.}
%Comparing Figure~\ref{fig:vis_entro} (a) and (c), NoPE shows a weak advantage of length generalization compared to RoPE, with the inflection point for NoPE is near position 2400, while RoPE is near position 2050. 
\textbf{the inflection point of $\widebar{\mathcal{H}}_i$ is highly consistent with the inflection point of perplexity}.
It implies that failed length generalization of NoPE (and RoPE)
might be connected to the distraction of attention:
attention heads begin to allocate attention to more tokens.
To further verify the connection,
we also draw a successful extension algorithm for RoPE 
(RoPE-NTK \cite{bloc97} which interpolates out-of-domain
encodings to in-domain encodings).
Its length generalization loss curve is flat,
while its entropy curve also has no steeply increasing point.

Unlike explicit position encodings, NoPE has no clear target objects
to manipulate, thus it is quite challenging to perform
length generalization without fine-tuning on longer sequences.
However, the strong correlation between length extension 
and attention pattern transition suggests such an object,
the entropy of attention heads.
%to improve NoPE's length generalization:

%confirming the correlation between the visualization metric and successful generalization.
%Moreover, the converged entropy for both NoPE and RoPE are respectively close to 6.0 and 6.2.

%\textbf{Observation 2: length generalization methods turn the steeply increasing $\widebar{\mathcal{H}}_i$ back to smooth.}
%Figure~\ref{fig:vis_entro} (b)  shows the NoPE length generalization method subsequently introduced in Section~\ref{ssec:uni_scale}, and (d) shows the RoPE length generalization method NTK.
%Despite the significant differences between these two methods, they both exhibit a similar change in entropy—transforming the steep increase back to smoothness.

%Based on the observations above, we find that a primary phenomenon of length generalization failure is distracted attention.
%While the RoPE can reconcentrate attention through modifications such as manipulating PE, the question arises whether there exists a method for NoPE to reconcentrate attention as well.
% refocus or reconcentrate

\subsection{Uniform Attention Scale}
\label{ssec:uni_scale}

We write the general scaled dot-product attention as
\begin{equation}
\alpha_{ij}^{(h)} = \frac{e^{\lambda \bm{q}_i^{(h)} \cdot \bm{k}_j^{(h)}}}{\sum_k e^{\lambda \bm{q}_i^{(h)} \cdot \bm{k}_k^{(h)}}}
\end{equation}
where the scaling factor 
$\lambda$ is the temperature hyper-parameter of the $\mathrm{SoftMax}$ operator.
The prevalent setting is $\lambda = \frac{1}{\sqrt{d}}$.
%(to boost training convergence).
%If $\lambda$ is set to 0, the scaled attention distribution becomes completely distracted, equivalent to a uniform distribution. If $\lambda$ is set very large, the scaled attention distribution becomes more concentrated, approaching a one-hot distribution.

Based on observations in Section~\ref{ssec:vis_entro}, 
we know that NoPE's failure of length generalization 
might be correlated with distracted attention, 
hence we can try to gradually increase the scale factor 
$\lambda$ to reconcentrate attention,
and see whether the generalization error can be reduced.
Figure~\ref{fig:vis_uni_entro} 
visualizes the average entropy under different scale values and the corresponding perplexity curves.

We first find that when increasing the scale factor 
during length generalization evaluation
(e.g., the pre-training scale $\lambda\!=\!\frac{1}{\sqrt{d}}$ 
is increased to $\lambda\!=\!\frac{1.2}{\sqrt{d}}$),
the inflection points of entropy curves are shifted to
longer lengths, at the same time, 
NoPE all generalize to further positions ($L \text{=2k}\to L^\prime\text{=4k}$). 
That is, with all NoPE's parameters frozen and 
only \emph{uniformly} increasing the softmax's temperature, 
NoPE can successfully generalize to unseen lengths.

The same conclusion doesn't hold for RoPE
(Figure~\ref{fig:vis_uni_entro} Right):
no matter what value the scale takes (from $\lambda$=0.8 to $\lambda$=1.4), 
the inflection points of entropy curves remain almost unchanged,
meaning that it fails to generalize to longer lengths.
On the other side, successful RoPE extension algorithms
(e.g., RoPE-NTK in Figure \ref{fig:vis_entro})
can control the distraction of entropy by
explicitly manipulate position encodings.
Therefore, though attention scaling has been used for RoPE
\cite{kexuefm8823,chiang2022overcoming},
it may contribute marginally to RoPE's generation.
%We can also hypothesis that, 

%\textbf{Finding 1: attention scaling enables successful generalization for NoPE but not for RoPE.}
%In Figure~\ref{fig:vis_uni_entro}, when a larger scale for length generalization evaluation, e.g., $\lambda\!=\!\frac{1.2}{\sqrt{d}}$, replaces the pre-training scale $\lambda\!=\!\frac{1}{\sqrt{d}}$, NoPE can generalize to farther positions ($L \text{=2k}\to L^\prime\text{=4k}$). 
%At the same time, the position where the average entropy appears to increase steeply is successfully backward-shifted by a larger scale factor.
%It indicates that by using just a single fixed scaling factor, NoPE can successfully generalize to unseen lengths.
%But for RoPE, no matter what value the scale takes (from $\lambda$=0.9 to $\lambda$=1.2), it fails to generalize to greater distances.
%While the convergence range of entropy may fluctuate up and down due to the scale's influence, the position where the entropy increases steeply remains essentially unchanged. 
%This is caused by the out-of-distribution (OOD) issue of explicit PE.
%Therefore, RoPE's generalization needs to address both the OOD and attention distraction issues simultaneously.
%\citet{kexuefm8823} and \citet{chiang2022overcoming} proposed that the use of an attention scaling factor enables better length generalization and does not require any modifications to the explicit PE.
%Attention scale is commonly used as an enhancement or supplementary method for RoPE generalization, and we are the \emph{first} to investigate its effectiveness in NoPE generalization.

We also find that 
extending NoPE to more distant positions generally requires a larger scale (i.e., a more concentrated attention distribution).
As the position becomes further, the number of tokens involved in the attention calculation increases, the attention is more easily scattered, 
and therefore, a larger scaling factor is needed to concentrate the attention.
In particular, for our NoPE model, generalizing to twice the pre-training length requires about 1.2 times the scale, four times the length requires about 1.5 times the scale, and eight times the length requires about 1.8 times the scale.
Appendix~\ref{app:fit_func} reports the fitted function of the scaling factor with respect to the generalization length $L'$.

%\textbf{Finding 2: generalizing to more distant positions requires a larger scale, i.e., a more concentrated attention distribution.}
%We also find that the position of successful NoPE generalization becomes larger with an increasing scale.
%As the position becomes larger, the number of tokens involved in the attention calculation increases, and at this point, the attention is more easily distracted. 
%Therefore a larger scaling factor is needed to reconcentrate the attention.
%\todo{values}
%Generalizing to twice the pre-training length requires about X times the scale, four times the length requires about Y times the scale, and eight times the length requires about Z times the scale.
%Appendix~\ref{app:fit_func} reports the fitted function of the scaling factor with respect to the generalization distance.

Finally, we note remark that the attention scaling factor in this section takes the \emph{same} value for all positions, including the pre-training length (\emph{uniform} scaling). 
We experimented with a piecewise function
which use the original scale within the pre-training positions, and a more concentrated attention scale for the extrapolated positions. 
We also try position-dependent functions, 
where the scale increases with position. 
However, none of these methods could further improve generalization. 
We speculate that if the attention at earlier positions is not highly concentrated, the learned token representations may hinder the concentration of attention at latter positions. 
We leave a deeper discussion and analysis of this observation in future work.

\input{fig/fig_vis_head_entro}

\section{Head-based Attention Scale}
\label{sec:head_scale}

After verifying that the attention scaling can 
help NoPE generalizing,
we delved deeper into the multi-head attention mechanism and posed a new question, ``\emph{Does each attention head require a unique scaling factor?}''

In this section, we first visualize the average entropy curves for each head and find that they have different attention patterns. 
Hence we propose to replace the uniform scaling with head-based scaling (from one factor to $22\times 32=704$ factors).
To address the issue of an exploding search space, we efficiently determine the values of scaling factors through automated hyperparameter search, considering both parameter efficiency and data efficiency.
As a result, head-based scaling generalizes better than uniform scaling.
Moreover, correlation analysis shows that within each layer, the smaller the converged entropy (i.e., the more concentrated attention), the larger the required scaling factor to maintain that concentration.

\subsection{Visual Analysis}
\label{ssec:vis_head_scale}

The entropy values span a broad spectrum, with each attention head demonstrating a distinct attention pattern. In Figure~\ref{fig:vis_head_entro}, certain attention heads show a highly concentrated pattern, with entropy values converging to $\approx 1$, while others exhibit a highly dispersed pattern, with entropy values converging to $\approx 10$. 
The full head visualization of Figure~\ref{fig:vis_head_entro} is located in Appendix~\ref{app:vis_all_heads}.

This phenomenon casts doubt on uniform scaling — how can a single scaling factor cater to diverse attention heads? 
Inspired by this, we further propose a head-based scale method.

\input{fig/fig_vis_head_vs_uni_scale}

\subsection{Head-based Scale}
\label{ssec:head_scale}

We reformulate the uniform attention scale as head-base attention scales
\begin{equation}
\alpha_{ij}^{(h)} = \frac{e^{\lambda^{(h)} \bm{q}_i^{(h)} \cdot \bm{k}_j^{(h)}}}{\sum_k e^{\lambda^{(h)} \bm{q}_i^{(h)} \cdot \bm{k}_k^{(h)}}}
\label{eq:head_scale}
\end{equation}
where $\lambda^{(h)}$ is a unique attention scaling factor for each head, totaling 704.
Compared to a uniform attention scale, 704 head-based scales make it difficult to determine the optimal values by grid search.
Similar to AutoML \citep{he2021automl}, we model the scales' optimal search as a parameter-efficient fine-tuning task.
Given a NoPE model $\mathcal{M}$ and a set of head-based scales $\{\lambda^{(1)}, \lambda^{(2)}, \dots, \lambda^{(m)}\}$, we fix the model $\mathcal{M}$ and define the head-based scales as trainable parameters $\theta=\{\lambda^{(1)}, \lambda^{(2)}, \dots, \lambda^{(m)}\}$.
We aim to find an optimal set of values $\theta^*=\{\lambda^{*(1)}, \lambda^{*(2)}, \dots, \lambda^{*(m)}\}$, that allows the model $\mathcal{M}{(\theta^*)}$ to successfully extend to the target length $L^\prime$.
To this end, we optimize the language modeling loss function $\mathcal{L}_{\text{LM}}$ on the pre-training dataset $D$ with length $L^\prime$ and size $n^\prime, n^\prime \ll n$.
\begin{equation}
\theta^* = \underset{x \in D}{\text{minimize}} \quad \mathcal{L}_{\text{LM}}\left(\mathcal{M}{(\theta, x)}\right)
\label{eq:head_scale_obj}
\end{equation}
The search process is highly efficient. (1) The number of tunable parameters is extremely small, only 704 delta parameters over 1B model parameters; 2) The amount of training tokens for fine-tuning is extremely small too, only 0.03\% of the pre-training data.

In addition, to ensure that the attention is reconcentrated instead of distracted 
by the scaling factors, we apply a focus constraint
during the optimization of Equation~\ref{eq:head_scale_obj} 
\begin{equation}
\lambda^{(*)} \ge \frac{1}{\sqrt d}
\label{eq:concentration}
\end{equation}

\input{fig/fig_vis_corr}

\paragraph{Initializing HeadScale}
In practice, we found that the initial value of head-based scales has a significant impact on the search of $\theta^*$.
An obvious approach is to use the default value $\lambda^{(*)}\!=\!\frac{1}{\sqrt{d}}$ from the pre-training phase. 
However, its length generalization results are quite unstable, with most being subpar,
as the optimal scale often deviates significantly from the default value.
We propose another approach to utilize the best uniform scale from the grid search as the initial value.
The ablation study for the initialization approach is in Section~\ref{ssec:ablation}.

Figure~\ref{fig:vis_head_vs_uni} compares the two generalization methods of NoPE, uniform scale versus head-based scales.
Head-based scale exhibits better generalization than the uniform scale, achieving a lower log-PPL by 0.2 at 4K positions ($2\!\times\!L$) and by 0.8 at 8K positions ($4\!\times\!L$).
The average entropy $\widebar{\mathcal{H}}_i$ of the head-based scale is higher than that of the uniform scale, suggesting that the uniform scale method over-concentrates attention, particularly for some heads that inherently have more distracted patterns.

Figure~\ref{fig:vis_corr} shows the correlation between the converged entropy and the searched scale.
To save space, we uniformly sampled 7 layers and all their respective heads. 
We observed that the correlation is layer-dependent, within each layer, heads with more concentrated attention (i.e., lower entropy) searched for larger scales, while heads with more dispersed attention (i.e., higher entropy) searched for smaller scales.
The result is as expected, the more concentrated the attention pattern, the larger the scaling factor needed to maintain its focus.
Furthermore, we observed that attention heads in lower layers are generally more dispersed, whereas heads in higher layers are generally more concentrated (note that this is not strictly observed).

%% file: fig/fig_vis_head_entro.tex
\begin{figure}[t]
  \centering
  \includegraphics[width=\linewidth]{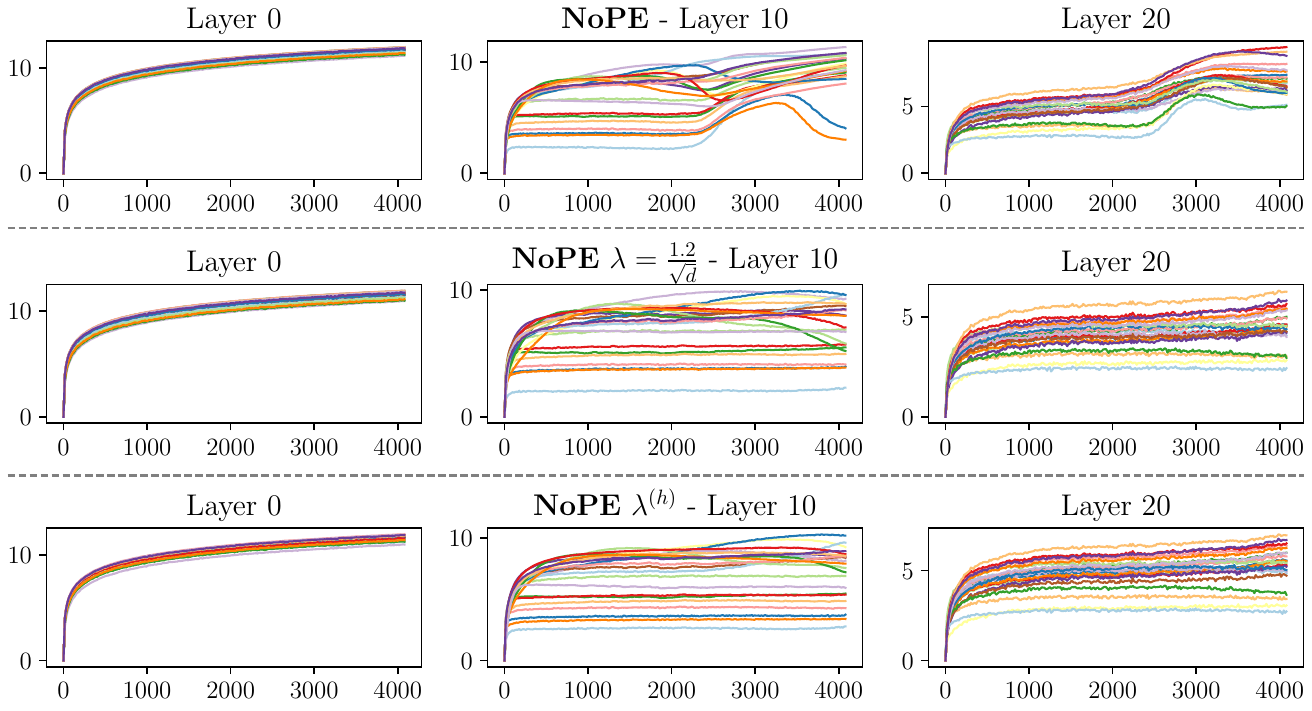}
  \caption{The attention entropy across all heads for the original NoPE, head-based scaled NoPE and uniform-scaled NoPE, with each model represented in a separate row. The attention heads exhibit divergent patterns.}
  \label{fig:vis_head_entro}
\end{figure}

%% file: fig/fig_vis_head_vs_uni_scale.tex
\begin{figure}[t]
  \centering
  \includegraphics[width=\linewidth]{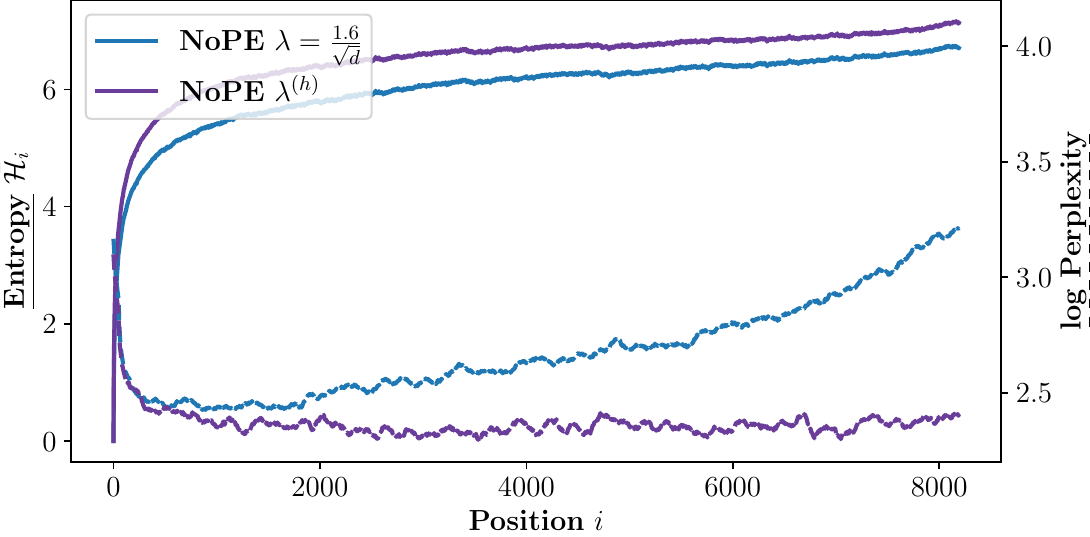}
  \caption{Comparing uniform and head-based scale (denoted as $\lambda^{(h)}$). UniformScale fails eventually as the perplexity increases with longer sequences. HeadScale is capable of handling much longer context by assigning different scale factors to each attention head.}
  \label{fig:vis_head_vs_uni}
\end{figure}

%% file: fig/fig_vis_corr.tex
\begin{figure}[t]
  \centering
  \includegraphics[width=\linewidth]{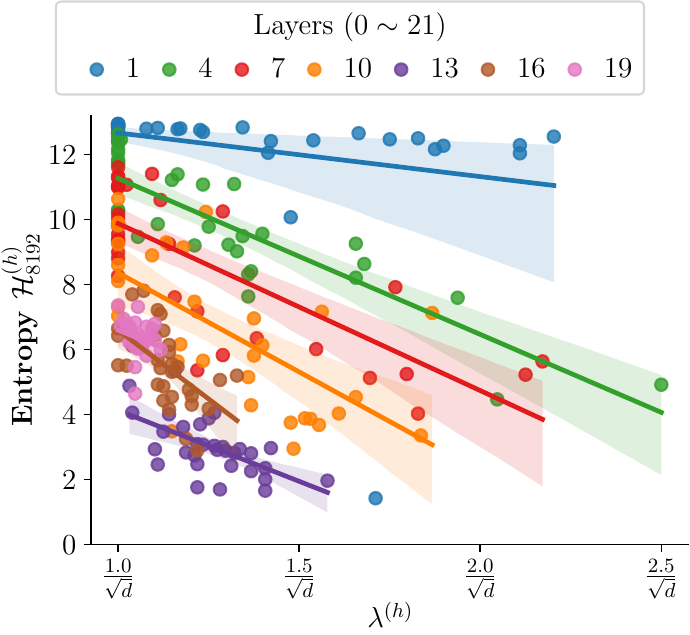}
  \caption{Correlation analysis for head-based scale when extended to 8K context. The analysis was conducted on the converged entropy values at 8K position, in relation to the scale searched. Each data point represents a unique attention head.}
  % \vspace{-0.5cm}
  \label{fig:vis_corr}
\end{figure}

%% file: sec/4_experiment.tex
\section{Experiment}
\label{sec:exper}

\input{tab/cs}

We train a NoPE base model from scratch and investigate its capability in length generalization. We conduct length generalization experiments on long sequence language modeling, synthetic tasks (passkey retrieval), and real-world long context tasks (LongBench). Detailed experiment setup can be found in Appendix~\ref{app:setup}.
% The results show that NoPE enjoys competitive performance compared to state-of-the-art length generalization methods for explicit position encodings (e.g., PI \cite{chen2023extending}, YaRN \cite{peng2024yarn}).

\input{tab/ppl}

\subsection{NoPE pre-trained model}
\label{ssec:nope}

For a fair comparison with RoPE, we train a NoPE model with $1.1$B parameters from the TinyLlama \cite{zhang2024tinyllama} code base\footnote{\url{https://github.com/jzhang38/TinyLlama}}. The NoPE model has $22$ layers of Transformer blocks, $32$ attention heads per layer, $2048$ embedding size. The model is trained on Slimpajama \cite{cerebras2023slimpajama} joint with Starcoderdata \cite{li2023starcoder} by $50$K steps ($\approx 100$B tokens) with sequence length $L=2048$. 

All settings are kept identical to those of TinyLlama, including the model architecture, training data, training procedure, and hyper-parameters, except that the rotary position embedding (RoPE) in TinyLlama is removed, making it a NoPE model, and the learning rate is set to $3.5\times10^{-4}$.

Following TinyLlama, we evaluate the commonsense reasoning ability of the NoPE model and report acc\_norm in Table~\ref{tab:cs}. We compare with the TinyLlama checkpoint that is trained on 100B tokens. The purpose of this experiment is to prove the NoPE base model performs on par with RoPE.

\subsection{Long Sequence Language Modeling}
\label{ssec:ppl}

Success on long sequence language modeling tasks is essential for length generalization.
A method that does not perform well in language modeling probably won't handle real-world long-context tasks.
% Any model that fails to achieve reasonable performance on language modeling tasks is unlikely to generalize well to longer sequences and accomplish real-world long-context tasks.

\paragraph{Settings.}
To evaluate the long sequence language modeling performances, we test our NoPE-based methods and RoPE-based baselines on PG19 \cite{Rae_Potapenko_Jayakumar_Hillier_Lillicrap_2020} and proof-pile \cite{proofpile} datasets.
For each dataset, we sample a subset of the test set and evaluate on $2$M tokens using sliding window evaluation ($S=256$) suggested by \citet{press2022train}.
We report the perplexity (PPL) of the models in Table~\ref{tab:ppl}.

\input{fig/fig_passkey.tex}

\paragraph{Main results.} 
Firstly, by comparing the original language models, NoPE's perplexity (PPL) is comparable to RoPE's for lengths within the training distribution, confirming the findings of \citet{haviv-etal-2022-transformer,chi-etal-2023-latent}. 
However, all LMs, including ALiBi models, fail to generalize out-of-the-distribution, indicating that explicit positional encoding is not the main reason for their failure in generalization. 
Current work on length generalization still focuses mainly on manipulating positional encoding.
Therefore, the length generalization issue within causal Transformer networks warrants a reanalysis and reinterpretation.

Secondly, by comparing the two generalization methods for NoPE proposed in this paper, the uniform scale method has significant limitations. 
Although using a larger scale can reduce the PPL at greater positions, it significantly affects the PPL at closer ranges. 
For instance, with a scale value of 1.8, the PPL on 2K@PG19 rises from 14.6 to 30.4, and on 2K@Proof-pile, it rises from 3.5 to 5.1.
On the contrary, the head-based scale method not only successfully extrapolates to 16k but also has minimal impact on the PPL at closer distances (for 18K, increases only +3.7 on 2K@PG19, +0.5 on 2K@Proof-pile), proving that attention heads with different patterns indeed require distinct scale values.

Third, a full comparison with RoPE LM's generalization method. 
Comparing the \emph{zero-shot} generalization methods, the head-based scale has better generalization than NTK, but weaker than YaRN.
In a fair comparison with the RoPE generalization methods which require \emph{fine-tuning}, 
the head-based scale method is competitive with these RoPE baselines, especially the Proof-pile dataset.
However RoPE baselines (PI, YaRN) still benefit from more training tokens, and the head-based scale on NoPE reaches its upper limit.
% First, comparing the original LM, NoPE trained with $50$K steps achieves competitive performance compared to the RoPE counterpart, which shows that NoPE is a valid approach to training large-scale casual language models.

% In general, head-based scale methods succeed in reducing PPL to the expected length by only adjusting 704 parameters.
% Head-based scale is competitive with RoPE baselines when fine-tuning with the same compute budget. While RoPE baselines (PI, YaRN) still benefit from more training tokens, the head-based scale on NoPE reaches its upper limit.

% However, in contrast to RoPE models, which achieve better perplexity when evaluating context window increases, the perplexity of NoPE models with head-based scale is observed to slowly degrade in most cases.
In summary, the head-based scale generalization method for NoPE slightly outperforms RoPE's early generalization method NTK, but still lags behind the recently introduced YaRN, particularly in near-distance PPL performance. 
Considering the significant challenge of generalizing NoPE compared to RoPE (due to the lack of explicit positional encoding to manipulate), this work, as the first to tackle length generalization for NoPE, has achieved its set goals.

The observed gap may imply that constraining the NoPE model to focus on fewer tokens could detrimentally affect its efficacy. Future efforts will be directed at enhancing the head-based scaling method to regain the level of performance seen in pretraining.

% We assume this indicates that forcing the NoPE model to concentrate on fewer tokens would hurt its performance eventually.
% More should be done to improve the head-based scale method to restore pretraining performance, and we leave that for further work.

\input{tab/long_bench}

\subsection{Synthetic Long Context Tasks}
\label{ssec:passkey}

A synthetic task is constructed in Landmark Attention \cite{mohtashami2023randomaccess} called "Passkey Retrieval".
It aims to test the effective context window size of the model.
The task is to retrieve a randomly placed passkey from a long sequence of tokens, where the passkey is a randomly sampled number of 5 digits and the sequence is built by concatenating irrelevant sentences.

\paragraph{Settings.} We evaluate the performance of passkey retrieval across various context lengths. For each specified context length, we conduct tests on 10 distinct passkey depths, each associated with 10 randomly selected passkeys. We report the retrieval accuracy in this task.

It is observed in Figure~\ref{fig:passkey} that both the NoPE base model and head-based scale perform well even when evaluating on $2\times$ the pretraining or fine-tuning context window, while RoPE strictly operates within the pre-trained sequence length and immediately fails outside of it. The result indicates that NoPE possesses significant potential for generalization.

\input{tab/ablation}

\subsection{Real-World Long Context Tasks}
\label{ssec:longbench}

LongBench \cite{bai2023longbench} is a comprehensive assessment of the long context understanding capabilities of large language models. We test all models using beam search decoding with beam size $5$. The evaluation context size is set to the model context window accordingly in order to test the model's capability to utilize a longer context. We only include raw PI and YaRN as the baseline in this task.

We find that the performance of the NoPE base model is better than its RoPE counterpart. Concluding better information utilization in the original length. Moreover, the head-based scale at a 4k extension length performs the best among all baselines. We attribute it to the capability of the NoPE base model and the successful length generalization of the head-based attention scale method.
While the head-based model still suffers from performance degradation when extending to a longer context, as it is stated in Section~\ref{ssec:ppl}.

\subsection{Ablation Study}
\label{ssec:ablation}

% Table~\ref{tab:ppl} shows PPL of uniform scale and head-based scale, as well as two variants of head-based scale: the one initialized with default value $\lambda^{(*)}=\frac{1}{\sqrt{d}}$ and the one with ReLU0.

% We observe that uniform scale performs well at $4$K length with a uniform attention scale $\lambda=\frac{1.2}{\sqrt{d}}$, but further increasing the uniform $\lambda$ does not flatten the PPL, on the contrary, it increases the PPL at $2$K length drastically.

We have introduced two key components of HeadScale in Section~\ref{ssec:head_scale}, a concentration constraint and an initializing technique. The ablation study in Table~\ref{tab:ablation} depicts that although occasionally perform better in language modeling, the two variants are less preferment in passkey retrieval and LongBench, indicating their inability to utilize long context information. 

Detailed results of the passkey retrieval task can be found in Figure~\ref{fig:ablation_pk} in the Appendix~\ref{app:abl_hs_pk}. They are completely unable to answer the passkey except when it is at the beginning of the context window.

%% file: tab/cs.tex
\begin{table*}[t]
    \centering
    \small
        \begin{tabular}{lccccccccc}
            \toprule
            \textbf{Model} & \textbf{Avg.} & \textbf{arc\_challenge} & \textbf{arc\_easy} & \textbf{boolq} & \textbf{hellaswag} & \textbf{openbookqa} & \textbf{piqa} & \textbf{winogrande}\\
            \midrule
            RoPE & 46.1 & \textbf{24.3} & \textbf{44.9} & \textbf{59.7} & \textbf{43.5} & 29.8 & 67.3 & \textbf{53.3} \\ 
            NoPE & \textbf{46.2} & 24.0 & \textbf{44.9} & 58.1 & 43.4 & \textbf{31.8} & \textbf{68.4} & 52.9 \\
            \bottomrule
        \end{tabular}
    \caption{Commonsense reasoning ability of the pre-trained base models. }
    \label{tab:cs}
\end{table*}

%% file: tab/ppl.tex
\begin{table*}[t]
    \centering
    \small
    \begin{tabular}{lrrrrrrrrrr}
    \toprule
        \multirow{2}{*}{\textbf{Model}} & \multicolumn{2}{c}{\textbf{FT}}   & \multicolumn{4}{c}{\textbf{PG19}} & \multicolumn{4}{c}{\textbf{Proof-pile}}\\
        \cmidrule(lr){2-3} \cmidrule(lr){4-7} \cmidrule(lr){8-11}
        & \bf $L^{\prime}$ & \bf Tokens & \textbf{2K} & \textbf{4K} & \textbf{8K} & \textbf{16K} & \textbf{2K} & \textbf{4K} & \textbf{8K} & \textbf{16K} \\ 
        \midrule
        \rowcolor{gray!10} \multicolumn{11}{c}{\textit{\textbf{Original LMs}}} \\
        RoPE & - & - & 14.5 & 491.4 & 488.5 & 599.5 & 3.5 & 303.0 & 432.1 & 759.5 \\ 
        NoPE & - & - & 14.6 & 326.9 & $>10^3$ & $>10^3$ & 3.5 & 117.4 & $>10^3$ & $>10^3$ \\
        BLOOM & - & - & 27.7 & 158.0 & 264.6 & 403.4 & 6.9 & 74.1 & 176.2 & 334.5\\
        MPT & - & - & 10.6 & 103.6 & 361.6 & 345.1 & 2.8 & 70.1 & $>10^3$ & $>10^3$ \\
        \arrayrulecolor{gray!20}
        \rowcolor{gray!10} \multicolumn{11}{c}{\textit{\textbf{Generalization for RoPE}}} \\
        NTK$^{\text{zero}}$ & - & - & 14.5 & 14.9 & 22.8 & 80.4 & 3.5 & 3.3 & 4.1 & 13.3 \\ 
        YaRN$^{\text{zero}}$ & - & - & 14.5 & 14.5 & 15.0 & 17.1 & 3.5 & 3.3 & 3.2 & 3.6 \\ 
        % PI$^{\text{fair}}$ & 4k & 6M & 15.98 & 15.85 & 551.92 & 1143.71 & 3.76 & 3.44 & 307.86 & 633.8 \\ 
        % PI$^{\text{fair}}$ & 8k & 13M & 17.37 & 17.13 & 17.12 & 752.82 & 3.96 & 3.6 & 3.36 & 406.34 \\ 
        % PI$^{\text{fair}}$ & 16k & 30M & 18.71 & 18.41 & 18.25 & 18.24 & 4.27 & 3.88 & 3.58 & 3.61 \\ 
        \hline
        \multicolumn{1}{l}{\multirow{3}{*}{PI$^{\text{fair}}$}}  & 4K & 6M & 16.0 & 15.9 & 551.9 & $>10^3$ & 3.8 & 3.4 & 307.9 & 633.8 \\ 
        \multicolumn{1}{l}{} & 8K & 13M & 17.4 & 17.1 & 17.1 & 752.8 & 4.0 & 3.6 & 3.4 & 406.3 \\ 
        \multicolumn{1}{l}{} & 16K & 30M & 18.7 & 18.4 & 18.3 & 18.2 & 4.3 & 3.9 & 3.6 & 3.6 \\ 
       
        \hline
        \multicolumn{1}{l}{\multirow{3}{*}{YaRN$^{\text{fair}}$}}  & 4K & 6M & 15.5 & 15.4 & 545.2 & $>10^3$ & 3.7 & 3.4 & 351.5 & 698.2 \\ 
        \multicolumn{1}{l}{} & 8K & 13M & 15.7 & 15.4 & 15.5 & 794.6 & 3.8 & 3.4 & 3.2 & 492.8 \\ 
        \multicolumn{1}{l}{} & 16K & 30M & 15.9 & 15.6 & 15.4 & 15.5 & 3.8 & 3.5 & 3.2 & 3.2 \\ 
        % YaRN$^{\text{fair}}$ & 4k & 6M & 15.5 & 15.39 & 545.23 & 1313.67 & 3.73 & 3.43 & 351.52 & 698.17 \\ 
        % YaRN$^{\text{fair}}$ & 8k & 13M & 15.7 & 15.4 & 15.46 & 794.64 & 3.76 & 3.41 & 3.2 & 492.76 \\ 
        % YaRN$^{\text{fair}}$ & 16k & 30M & 15.91 & 15.64 & 15.42 & 15.52 & 3.81 & 3.46 & 3.18 & 3.23 \\ 
        % HS18k\_i1 & 18.21 & 18.78 & 19.02 & 31.37 & 3.93 & 3.62 & 3.31 & 4.3 \\ 
        % HS18k\_relu0 & 18.04 & 18.56 & 18.01 & 25.85 & 3.91 & 3.58 & 3.25 & 4.15 \\ 
        % PI$^{\text{raw}}$ & 4k & 33M & 15.17 & 15.04 & 623.79 & 951.73 & 3.6 & 3.29 & 334.47 & 595.54 \\ 
        % PI$^{\text{raw}}$ & 8k & 66M & 15.38 & 15.1 & 14.97 & 909.64 & 3.63 & 3.28 & 3.04 & 463.03 \\ 
        % PI$^{\text{raw}}$ & 16k & 131M & 15.63 & 15.31 & 15.02 & 14.78 & 3.67 & 3.32 & 3.01 & 2.98 \\ 
        \hline
        \multicolumn{1}{l}{\multirow{3}{*}{PI$^{\text{raw}}$}} & 4K & 33M & 15.2 & 15.0 & 623.8 & 951.7 & 3.6 & 3.3 & 334.4 & 595.5 \\ 
        \multicolumn{1}{l}{} & 8K & 66M & 15.4 & 15.1 & 15.0 & 909.6 & 3.6 & 3.3 & 3.0 & 463.0 \\ 
        \multicolumn{1}{l}{} & 16K & 131M & 15.6 & 15.3 & 15.0 & 14.9 & 3.7 & 3.3 & 3.0 & 3.0 \\ 
        % YaRN$^{\text{raw}}$ & 4k  & 33M & 15.1 & 15 & 573.32 & 951.37 & 3.62 & 3.33 & 358.84 & 656.81 \\ 
        % YaRN$^{\text{raw}}$ & 8k  & 66M & 15.09 & 14.83 & 14.8 & 815.97 & 3.61 & 3.27 & 3.05 & 501.51 \\ 
        % YaRN$^{\text{raw}}$ & 16k  & 131M & 15.04 & 14.75 & 14.52 & 14.5 & 3.6 & 3.25 & 2.99 & 2.99 \\ 
        \hline
        \multicolumn{1}{l}{\multirow{3}{*}{YaRN$^{\text{raw}}$}} & 4K  & 33M & 15.1 & 15.0 & 573.3 & 951.4 & 3.6 & 3.3 & 358.8 & 656.8 \\ 
        \multicolumn{1}{l}{}& 8K  & 66M & 15.1 & 14.8 & 14.8 & 816.0 & 3.6 & 3.3 & 3.1 & 501.5 \\ 
        \multicolumn{1}{l}{}& 16K  & 131M & 15.0 & 14.8 & 14.5 & 14.5 & 3.6 & 3.3 & 3.0 & 3.0 \\ 
        
        \rowcolor{gray!10} \multicolumn{11}{c}{\textit{\textbf{Generalization for NoPE}}} \\
        $\lambda\!=\!\frac{1.2}{\sqrt{d}}$  & - & -  & 15.0 & 16.0 & 513.7 & $>10^3$ & 3.6 & 3.3 & 175.3 & $>10^3$ \\ 
        $\lambda\!=\!\frac{1.5}{\sqrt{d}}$  & - & -  & 19.0 & 20.2 & 45.3 & 224.1 & 3.9 & 3.7 & 4.9 & 99.2 \\ 
        $\lambda\!=\!\frac{1.8}{\sqrt{d}}$  & - & -  & 30.4 & 42.4 & 69.1 & 198.8 & 5.1 & 5.6 & 8.5 & 38.2 \\ 
        % $\lambda^{(h)}$ & 4k & 6M & 14.76 & 15.32 & 404.5 & 3224.27 & 3.51 & 3.24 & 153.49 & 1676.57 \\ 
        % $\lambda^{(h)}$ & 8k & 13M & 15.71 & 15.32 & 21.14 & 721.72 & 3.61 & 3.25 & 3.21 & 318.46 \\ 
        % $\lambda^{(h)}$ & 18k & 30M & 18.31 & 18.99 & 18.83 & 30.42 & 3.98 & 3.68 & 3.32 & 4.08 \\ 
        \hline
        \multicolumn{1}{l}{\multirow{3}{*}{$\lambda^{(h)}$}} & 4K  & 6M & 14.8 & 15.3 & 404.5 & $>10^3$ & 3.5 & 3.2 & 153.4 & $>10^3$ \\ 
        \multicolumn{1}{l}{} & 8K & 13M & 15.7 & 15.3 & 21.1 & 721.7 & 3.6 & 3.3 & 3.2 & 318.5 \\ 
        \multicolumn{1}{l}{} & 18K & 30M & 18.3 & 19.0 & 18.8 & 30.4 & 4.0 & 3.7 & 3.3 & 4.1 \\ 
        \arrayrulecolor{black}
        \bottomrule
        \vspace{-0.5cm}
    \end{tabular}
    \caption{Sliding window perplexity of different context window extension methods tested on PG19 and ProofPile. The ``fair'' and ``raw'' versions of PI and YaRN differ from the training data, as detailed in Appendix~\ref{app:setup}. The notation $\lambda=*$ denotes uniform attention scale by the given number, and $\lambda^{(h)}$ represents head-based scale. }
    \label{tab:ppl}
\end{table*}

%% file: fig/fig_passkey.tex
\begin{figure*}[t]
  \centering
  \includegraphics[width=0.32\linewidth]{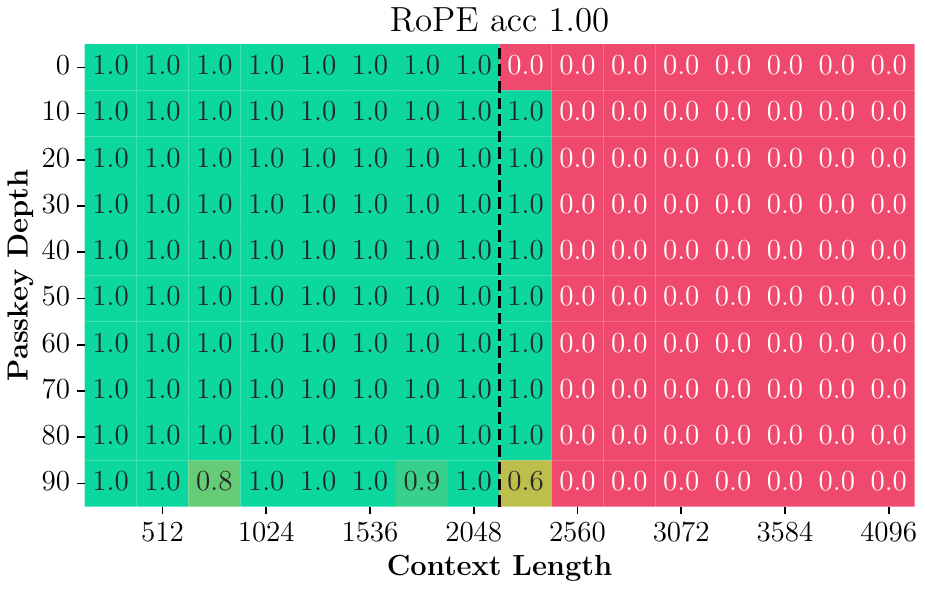}
  \includegraphics[width=0.32\linewidth]{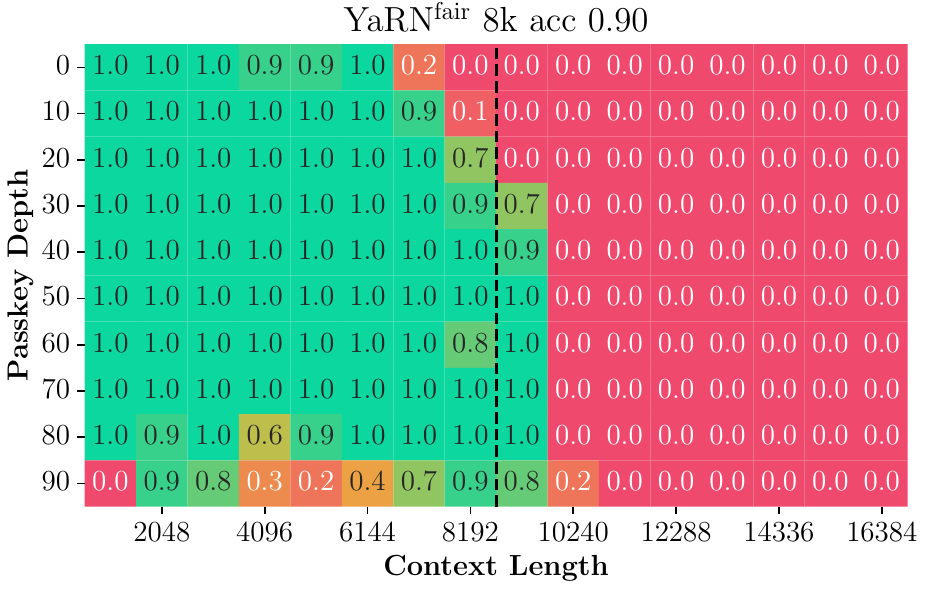}
  \includegraphics[width=0.32\linewidth]{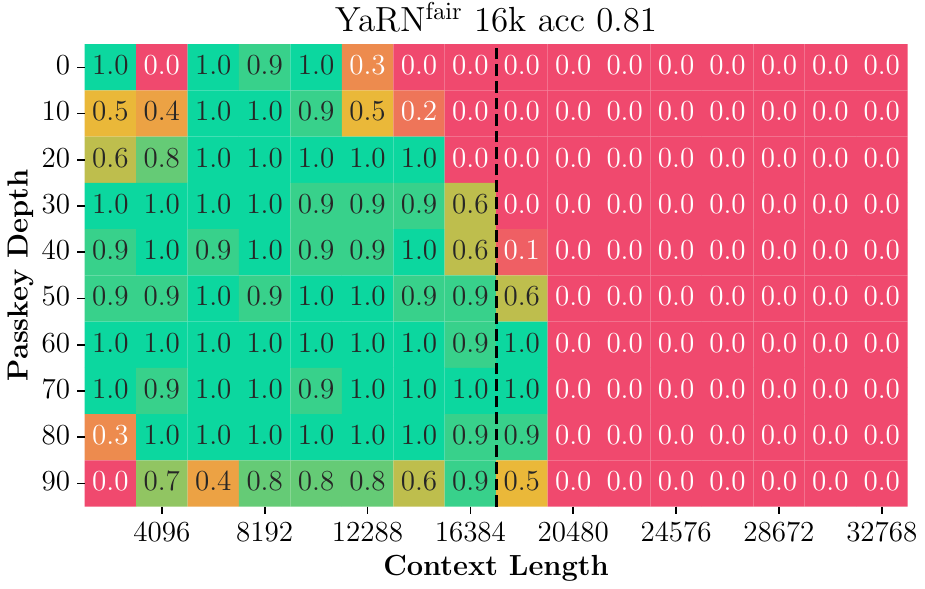}\\
  \includegraphics[width=0.32\linewidth]{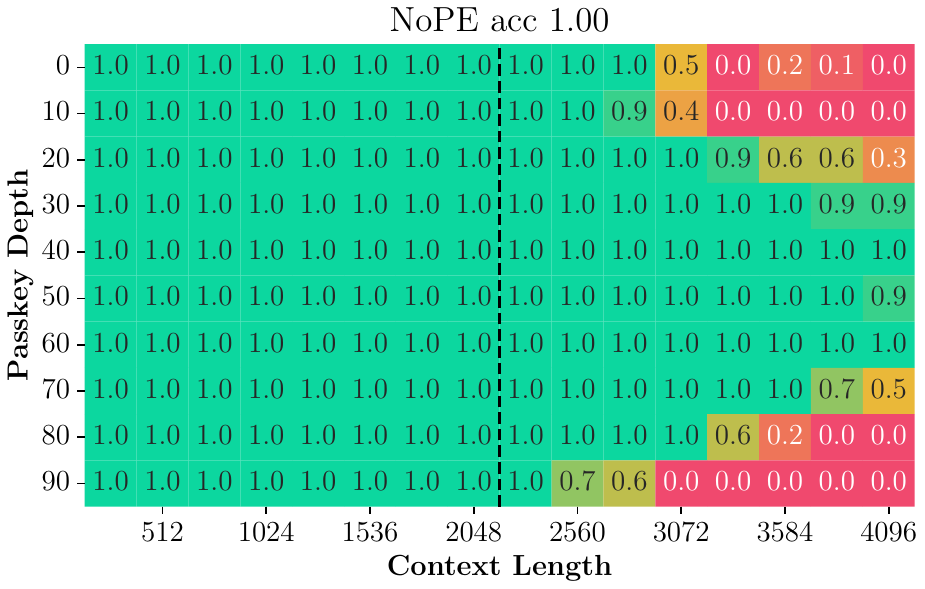}
  \includegraphics[width=0.32\linewidth]{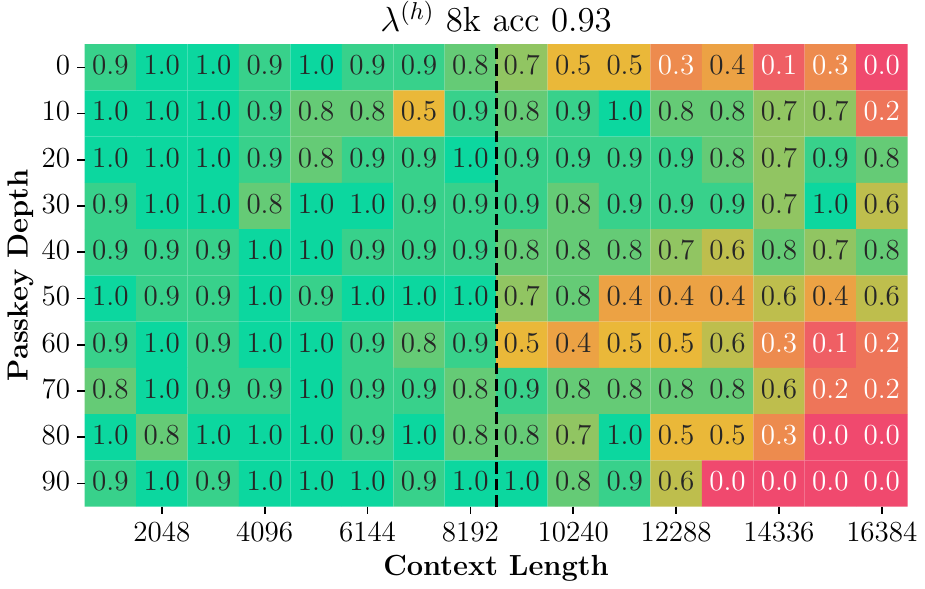}
  \includegraphics[width=0.32\linewidth]{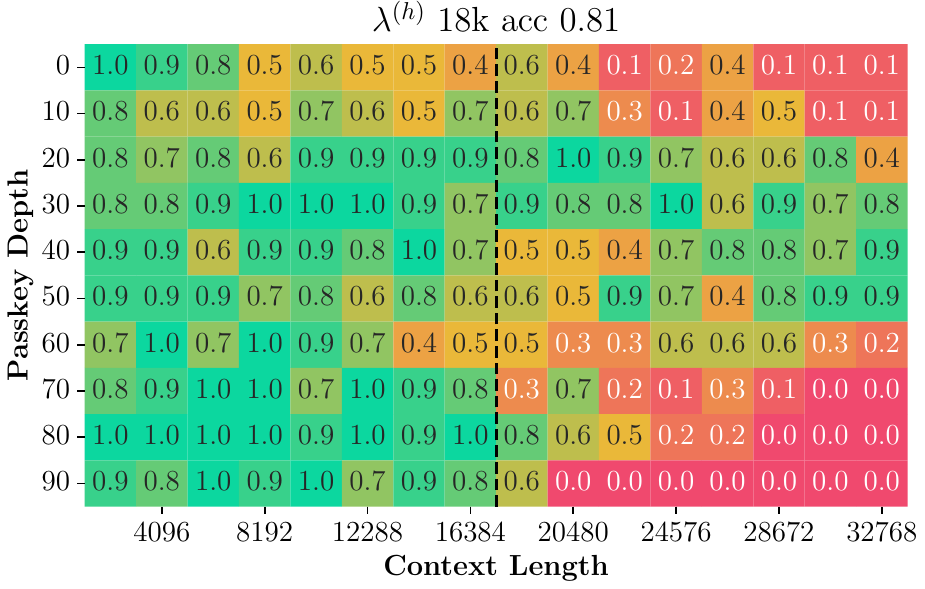}
  % \vspace{-0.2cm}
  \caption{The figures illustrate the passkey retrieval accuracy for both RoPE and NoPE methods. The vertical dashed line represents the context length of the models, which could be either the pre-training length or the fine-tuning length. The title of each sub-figure indicates the average accuracy within the model’s context length. Notably, NoPE demonstrates robust performance even beyond the model’s context window, indicating significant potential for generalization.}
  \label{fig:passkey}
\end{figure*}

%% file: tab/long_bench.tex
\begin{table*}[!t]
    \centering

    \resizebox{\textwidth}{!}{
        \begin{tabular}{lrrrrrrrrrrrrrrrrrr}
        \toprule
        \multirow{2}{*}{\textbf{Model}} & \multirow{2}{*}{\textbf{Ctx.}} & \multirow{2}{*}{\textbf{Avg.}} & \multicolumn{3}{c}{\textbf{Singl-Doc QA}} & \multicolumn{3}{c}{\textbf{Multi-Doc QA}} & \multicolumn{3}{c}{\textbf{Summarization}} & \multicolumn{3}{c}{\textbf{Few-shot Learning}} & \multicolumn{2}{c}{\textbf{Synthetic}} & \multicolumn{2}{c}{\textbf{Code}} \\
        \cmidrule(lr){4-6} \cmidrule(lr){7-9} \cmidrule(lr){10-12} \cmidrule(lr){13-15} \cmidrule(lr){16-17} \cmidrule(lr){18-19} 
        & & & \textbf{NQA} & \textbf{Qsp} & \textbf{MulF} & \textbf{HpQA} & \textbf{2WQA} & \textbf{Musq.} & \textbf{GRpt} & \textbf{QSum} & \textbf{MulN} & \textbf{TREC} & \textbf{TrQA} & \textbf{SSum} & \textbf{PsgC} & \textbf{PsgR} & \textbf{Lcc} & \textbf{Re-P} \\
        \midrule
        \rowcolor{gray!10} \multicolumn{19}{c}{\textit{\textbf{Original LMs}}} \\
        RoPE & 2K & 16.5 & 3.5 & 4.7 & 17.5 & 3.4 & 8.8 & 2.8 & 26.9 & 8.4 & \textbf{25.9} & 33.5 & 18.8 & 15.7 & 1.9 & 2.5 & 49.5 & 40.1 \\ 
        NoPE & 2K & 18.3 & 6.1 & 7.9 & 22.4 & 6.6 & 10.3 & 3.1 & \textbf{28.9} & 8.8 & 25.1 & \textbf{41.5} & 30.0 & 3.5 & 1.0 & 3.0 & 48.4 & 46.6 \\ 
        \arrayrulecolor{gray!20}
        \rowcolor{gray!10} \multicolumn{19}{c}{\textit{\textbf{Generalization for RoPE}}} \\
        \multicolumn{1}{l}{\multirow{3}{*}{PI$^{\text{raw}}$}} & 4K & 16.7 & 5.4 & 8.6 & 18.6 & 4.5 & 9.1 & 3.9 & 26.4 & 9.9 & 18.5 & 21.5 & 21.2 & 22.2 & \textbf{2.7} & 1.5 & 48.5 & 44.6 \\ 
        \multicolumn{1}{l}{} & 8K & 16.7 & 4.7 & 9.6 & 16.3 & 5.4 & 9.3 & 4.0 & 14.6 & 9.4 & 20.7 & 27.0 & 23.1 & 23.5 & 2.1 & 3.4 & 50.0 & 44.7 \\ 
        \multicolumn{1}{l}{} & 16K & 17.2 & 4.8 & 8.1 & 18.6 & 5.4 & 9.4 & 3.8 & 22.9 & 9.9 & 21.3 & 24.0 & 23.9 & \textbf{25.4} & 1.6 & 1.8 & \textbf{50.5} & 43.8 \\ 
        \hline
        \multicolumn{1}{l}{\multirow{3}{*}{YaRN$^{\text{raw}}$}} & 4K & 16.2 & \textbf{6.4} & 8.7 & 18.2 & 4.0 & 11.0 & 3.0 & 17.5 & 9.0 & 15.6 & 27.5 & 21.5 & 20.3 & 1.6 & 0.5 & 49.8 & 45.2 \\ 
        \multicolumn{1}{l}{\multirow{3}{*}{}} & 8K & 16.4 & 6.0 & 11.4 & 16.0 & 5.0 & 8.3 & 3.5 & 16.3 & \textbf{10.3} & 19.6 & 21.0 & 24.9 & 22.1 & 1.3 & 2.0 & 49.6 & 45.3 \\ 
        \multicolumn{1}{l}{\multirow{3}{*}{}} & 16K & 17.7 & 4.5 & 10.5 & 17.1 & 5.2 & 8.9 & \textbf{4.7} & 18.9 & 9.2 & 19.5 & 38.0 & 24.4 & 25.2 & 1.7 & 1.8 & 49.8 & 44.6 \\ 
        \rowcolor{gray!10} \multicolumn{19}{c}{\textit{\textbf{Generalization for NoPE}}} \\
        \multicolumn{1}{l}{\multirow{3}{*}{$\lambda^{(h)}$}} & 4K & \textbf{18.5} & 6.3 & 11.1 & \textbf{23.1} & 5.7 & 10.1 & 4.2 & 27.7 & 8.9 & 23.4 & 25.5 & \textbf{35.7} & 13.7 & 0.6 & \textbf{4.5} & 47.9 & \textbf{46.9} \\ 
        \multicolumn{1}{l}{\multirow{3}{*}{}} & 8K & 17.2 & 5.8 & 11.7 & 21.4 & 6.1 & 10.8 & 3.9 & 24.1 & 8.9 & 18.3 & 31.0 & 31.4 & 4.5 & 0.6 & 3.1 & 47.3 & 46.5 \\ 
        \multicolumn{1}{l}{\multirow{3}{*}{}} & 18K & 17.0 & 6.0 & \textbf{12.8} & 20.3 & \textbf{7.0} & \textbf{12.9} & 4.1 & 17.2 & 8.4 & 16.1 & 41.0 & 32.9 & 5.1 & 0.3 & 2.1 & 44.5 & 41.0 \\ 
        \arrayrulecolor{black}
        \bottomrule
        \end{tabular}
    }
    \caption{Real-world Long-Context performance of NoPE-extension methods and various RoPE baselines. The ``Ctx.'' column represents testing context length during evaluation, which corresponds to either the pre-training length for base models or the extended length for length generalization methods.}
    \label{tab:longbench}
\end{table*}

%% file: tab/ablation.tex
\begin{table}[t]
    \centering
    \resizebox{\linewidth}{!}{
        \begin{tabular}{lcccc}
        \toprule
        \multirow{2}{*}{\textbf{Model}} & \multicolumn{2}{c}{\textbf{PPL@16K ($\downarrow$)}}  & \multirow{2}{*}{\textbf{Passkey ($\uparrow$)}} & \multirow{2}{*}{\textbf{LongBench ($\uparrow$)}} \\
        \cmidrule(lr){2-3}
        & \textbf{PG19} & \textbf{Proof-pile} & &\\
        \midrule
        $\lambda^{(h)}$ 18K & 30.4 & \textbf{4.1} & \textbf{81} & \textbf{17.0} \\ 
        w/o focus constraint & \textbf{25.9} & 4.2 & 53 & 16.7 \\
        w/o initialization & 31.4 & 4.3 & 26 & 15.8 \\
        \bottomrule
        \end{tabular}
    }
    \caption{Ablation study on the two variants of HeadScale. Passkey results are listed as average accuracy, and LongBench results are averaged score among all sub-tasks.}
    \label{tab:ablation}
\end{table}

%% file: sec/5_related_work.tex
\section{Related Work}
\label{sec:related-work}

% \paragraph{Position encodings}.
% Absolute position encoding,
% Relative position encoding.

\paragraph{Transformers without position encoding}
% three NoPE papers.
\citet{haviv-etal-2022-transformer} was the first to discover that causal Transformer networks could perform language modeling tasks successfully even without explicit PE.
\citet{chi-etal-2023-latent} provided a theoretical explanation for NoPE, demonstrating that for an initialized NoPE LM, the variance of the hidden representations in each layer is position-dependent, with variance decreasing for larger positions.
Both works demonstrate that the NoPE hidden layer representation implies positional information through the probing task.
\citet{kazemnejad2023the} proved through constructive methods that NoPE can learn absolute PE from the first layer and relative PE from the second layer. 
They also showed that NoPE has an extremely weak length generalization ability (train $\sim$20, test $\sim$40), but is slightly better than LM with explicit PE.
This paper first proposes length generalization methods for NoPE with uniform scale and head-based scale.
For the first time verifies the effectiveness of NoPE generalization in real LLM settings.

\paragraph{Length generalization}
% Train short, test long (Alibi),
% position interpolation (PI),
% NTK, Yarn.
% Differences and connections with the method proposed here.
Due to high computational and memory requirements, LLM training is usually limited to short inputs. 
Directly applying LLMs to long inputs faces the challenge of out-of-distribution (OOD) issues. 
Research to enable LLMs to process long inputs has been extensive \cite{huang2023advancing,dong2023survey}. 
The earliest methods involved designing new relative PE mechanisms during pre-training
\citep{press2021train,sun-etal-2023-length}. 
Subsequent studies focused primarily on the widely used RoPE \cite{su2024roformer} and proposed length extension by mitigating RoPE's OOD issues through interpolated positions
\citep{chen2023extending,kaiokendev9444,peng2023yarn,emozillareddit,bloc97,bloc972}. 
Other works employed sliding window attention mechanisms to prevent relative positions from exceeding the maximum distance seen in pre-training 
\citep{mohtashami2023landmark,han2023lminfinite,xiao2023efficient,jin2024llm,zhang2024soaring}. 
However, these models ignore information from distant tokens, thus failing to capture long-distance context dependencies.
All existing methods rely on specific explicit PEs. 
However, the NoPE architecture is more streamlined and more aligned to the form of human language modeling.
Exploring NoPE's length generalization is therefore more intriguing and attractive.

%% file: sec/6_conclusion.tex
\section{Discussion}
We studied the length generalization of Casual Transformer 
without explicit position encoding.
We developed a parameter-efficient tuning algorithm 
which aims to search for the best temperature hyper-parameters
for attention heads.
Through empirical evaluation, we saw that 
NoPE can achieve competitive length generalization and might be 
a promising alternative for long-context language modeling.

NoPE provides a new perspective to understanding the role of positional information by isolating and eliminating the effects of explicit positional encoding. Our work demonstrates the correlation between length generation failures and distraction of attention in NoPE models, thus the proposed method concentrates the attention by adjusting the scaling factor. While current works on length generalization mainly focus on manipulating positional encoding, our work suggests a new key component to generalization.

\section*{Limitation}
The length generalization algorithms discussed in this paper exhibit 
 competitive performances, but the NoPE model itself still
 underperforms with state-of-the-art RoPE models, which 
 makes the results over long sequence language modeling tasks 
 and LongBench tasks are less competitive.
 NoPE still faces the challenges of considerable memory usage and computational complexity due to the quadratic nature of attention computation when processing extremely long contexts. 
Hardware limitations are likely to become a constraining factor for length generalization soon.
 We plan to further 
 improve the NoPE's performances for a fairer comparison.
 This paper is also most an empirical one, which requires 
 a deeper theoretical understanding of NoPE's length generalization
 in the future.

%Quadratic complexity of attention computation. 
%Although our proposed method enables NoPE to generalize successfully without explicit positional encoding, NoPE still faces the challenges of considerable memory usage and computational complexity due to the quadratic nature of attention computation when processing extremely long contexts. 
%Hardware limitations are likely to become a constraining factor for length generalization soon.

\section*{Acknowledgement}
The authors wish to thank all reviewers for their helpful comments and suggestions.
The corresponding authors are Tao Ji, Yuanbin Wu and Xiaoling Wang.
This research was (partially) supported by NSFC(62076097), National Key R\&D
Program of China (2021YFC3340700),  the Open Research Fund of Key Laboratory of Advanced Theory and Application in Statistics and Data Science (East China Normal University), Ministry of Education.

%% file: sec/8_appendix.tex
\section{Experiment Setup}
\label{app:setup}
% \subsection{Setup}
% \label{ssec:setup}

\paragraph{Searching scales.}
We approach the search for optimal head-based scales $\lambda^{(h)}$ by parameter-efficient fine-tuning.
We use a large learning rate (LR, =$0.05$ or =$0.1$) for fine-tuning, as $\lambda$ spans a wide range, (e.g., $[\frac1{\sqrt d}, \frac3{\sqrt d}]$, shown in Figure~\ref{fig:vis_corr}).
% We model the optimal search for scales as a parameter-efficient fine-tuning task, with only $704$ trainable parameters.
% Notice that $\lambda$ ends up being in a wide range (e.g., $[\frac1{\sqrt d}, \frac3{\sqrt d}]$, see Figure~\ref{fig:vis_corr}), so we need a large learning rate during fine-tuning.
% We set the learning rate to $0.1$ or $0.05$ and choose the best model.
The fine-tuning data comes from the pretraining dataset (Slimpajama \cite{cerebras2023slimpajama} and Starcoderdata \cite{li2023starcoder}) with a different data fetching seed from the pretraining.
We set the batch size to $8$ and set the optimizer to the AdamW ($\beta_1=0.9$, $\beta_2=0.95$) without weight decay \cite{Loshchilov2017DecoupledWD}.
We use a cosine LR decay from LR to $0.1$LR for $200$ fine-tuning steps and a linear warmup for the first $20$ steps.
% We fine-tuned for $200$ steps with a batch size of $8$ using the AdamW optimizer \cite{Loshchilov2017DecoupledWD} with $\beta_1=0.9$, $\beta_2=0.95$, and no.
We found that the head-based scale searching on $16$K suffers from a minor PPL degradation at the end of the context window.
We simply expanded the length $L^\prime$ to $18$K and then solved it.

\paragraph{Length generalization baselines.}
To compare with mainstream length generalization research, we reproduced three generalization baselines on RoPE, including:

\begin{itemize}[leftmargin=*,topsep=5pt,itemsep=0pt]
    \item NTK (\citeyear{blocntkaware}), zero-shot generalization; 
    \item PI (\citeyear{chen2023extending}), efficiently train long, test long;
    \item YaRN (\citeyear{peng2024yarn}), supports both settings 
    \footnote{
The YaRN paper also proposes a ``train short, test long'' setting with lower training costs. However, for a fair comparison, we relax this setting to ``train long, test long'' which generalizes better.}.
\end{itemize}

For the zero-shot setting, we grid-searched the baseline hyper-parameters and \textbf{reported their best results}.
For the baselines that need fine-tuning, we propose two settings, one for a fair comparison, with the same number of fine-tuned tokens (0.3\textperthousand~of pre-trained data) as the head-based scales searching, and the other follows their original paper, which is 1.3\textperthousand~of pre-trained data. Specifically, we fine-tune the RoPE model for 200 steps in the ``fair'' version, and 1000 steps for the ``raw'' version.
In addition, we incorporate open-source ALiBi models \citep{press2022train} into our baselines, which include BLOOM 1.1B \citep{bigscience_workshop_2022} and MPT 7B Base \citep{MosaicML2023Introducing}, both of which are trained on a context length of 2K. We test a zero-shot generalization of the ALiBi models following the original paper \citep{press2022train}.

\section{Fitted Function of the Uniform Scale}
\label{app:fit_func}

\input{fig/fig_fit}

In the study depicted in Figure~\ref{fig:fit}, a hyper-parameter search was conducted for the uniform scale $\lambda$ with an interval of $\frac{0.01}{\sqrt{d}}$.
This search was applied to two checkpoints of the pre-trained NoPE model, to fit the optimal $\lambda$ at the extension length. We note remark that the scaling factor takes the \emph{same} value for all positions during a single test. The output of a single test is the perplexity across all positions. We run multiple tests with different scales and find the best one for each position.

Based on the search results, we guess a function form that best fits the data points. We then fit this function over the range $i \in [2048,16384]$. The fitted function, along with its corresponding coefficient of determination, is presented below:

\begin{itemize}[leftmargin=*]
    \item For NoPE at 10k steps, the coefficient of determination $R^2=0.9954$. The fitted function is $$\lambda=\frac{1+0.3010\ln s}{\sqrt{d}}$$
    \item For NoPE at 50k steps, the coefficient of determination $R^2=0.9773$. The fitted function is $$\lambda=\frac{1+0.3973\ln s}{\sqrt{d}}$$
\end{itemize}

In these functions, $s$ is defined as $\frac iL$ for each position $i$, representing the model's extension ratio relative to its pre-training length.

Furthermore, it is also found by \citet{peng2024yarn} that the YaRN method benefits from a similar uniform scale on LLaMA2 \cite{touvron2023llama}, although the scale does not have a direct impact on the RoPE extension capability (refer to Figure~\ref{fig:vis_uni_entro}). The scale proposed by the YaRN method can be formulated as follows, which is quite similar to our result. $$\lambda=\frac{(1+0.1 \ln s)^2}{\sqrt{d}}$$ %\approx\frac{1+0.2 \ln s}{\sqrt{d}}$$

In conclusion, the optimal uniform scale varies across different models.
It is also observed from Figure~\ref{fig:fit} that uniform scale, despite being optimal, cannot flatten the NoPE model's perplexity within a large context window.
This finding underscores the importance of employing a head-based scaling method for managing model perplexity effectively across larger context windows, thereby enhancing the model’s performance.

\section{Additional Passkey Results}
\label{app:abl_hs_pk}

\input{fig/fig_pk_alibi}

In Section~\ref{ssec:ppl}, we note that the ALiBi baselines do not exhibit competitive performance in terms of perplexity when applied to longer contexts. We also conduct Passkey Retrieval tests on these models, with the results depicted in Figure~\ref{fig:pk_alibi}. These models yield expected results within their pre-trained sequence length, but they are unable to complete the task when it exceeds this length.

\input{fig/fig_ablation_pk}

In Section~\ref{ssec:ablation}, we conducted an ablation study on HeadScale. Figure~\ref{fig:ablation_pk} shows the passkey retrieval task of the two variations of HeadScale.

\section{Entropy Visualization of All Heads}
\label{app:vis_all_heads}

\Cref{fig:all_heads_8k,fig:all_heads_uni,fig:all_heads_ori} show attention entropy across all layers and all heads of the 8k extension head-based scale method, UniformScale and the original NoPE. 
An additional theoretical upper bound of entropy is also plotted in the figures. We note that for each position $i$, the maximum entropy is achieved when $\forall j,\  \alpha_{ij}^{(h)}=\frac 1i$ is satisfied in Equation~\ref{eq:entropy}. The maximum value is then given by $\mathcal{H}_i^{(h)}=\log i$.

\input{fig/fig_all_heads_ent_8k}
\input{fig/fig_all_heads_ent_uni}
\input{fig/fig_all_heads_ent_ori}

It is observed in Figure~\ref{fig:all_heads_8k} that the lower layers have high entropy, closely approaching the upper bound. Most heads exhibit constant entropy for all positions. And the attention values span a broad spectrum, ranging from 0 to theoretical upper-bound.

%% file: fig/fig_fit.tex
\begin{figure*}[t]
  \centering
  \includegraphics[width=0.49\linewidth]{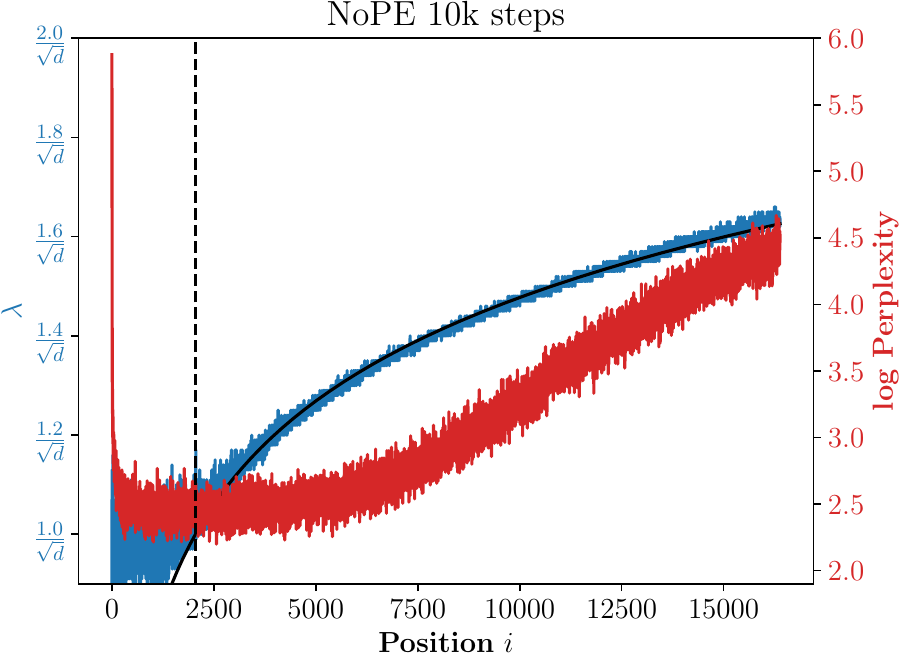}
  \includegraphics[width=0.49\linewidth]{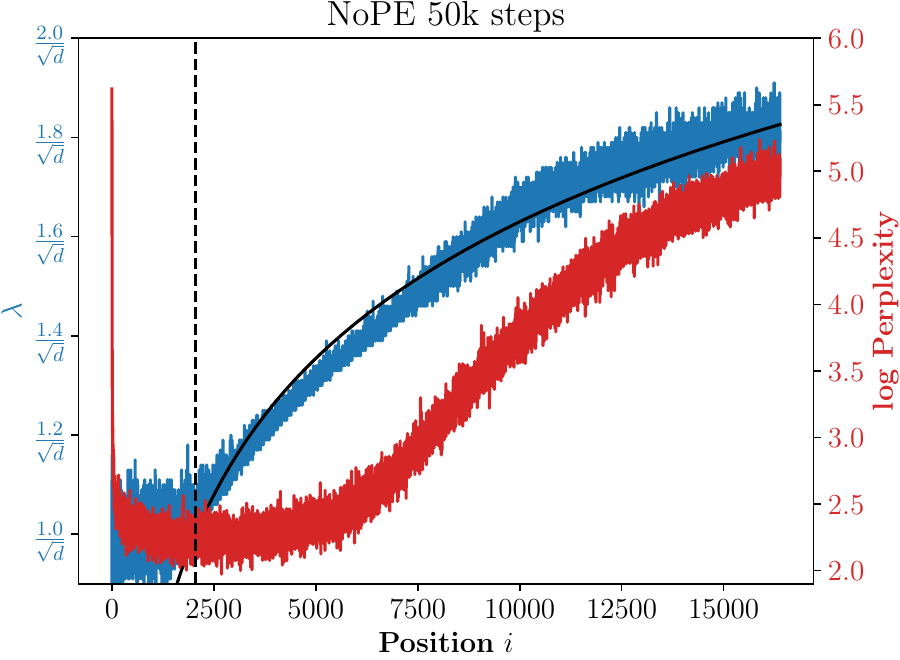}
  \caption{Fitted optimal uniform scale for each position. The red line indicates best log perplexity found at each position, the blue line plots the corresponding optimal uniform $\lambda$ for that position, the black curve is the fitted function and the vertical dotted line is pre-training length.}
  \label{fig:fit}
\end{figure*}

%% file: fig/fig_pk_alibi.tex
\begin{figure*}[t]
  \centering
  \includegraphics[width=0.49\linewidth]{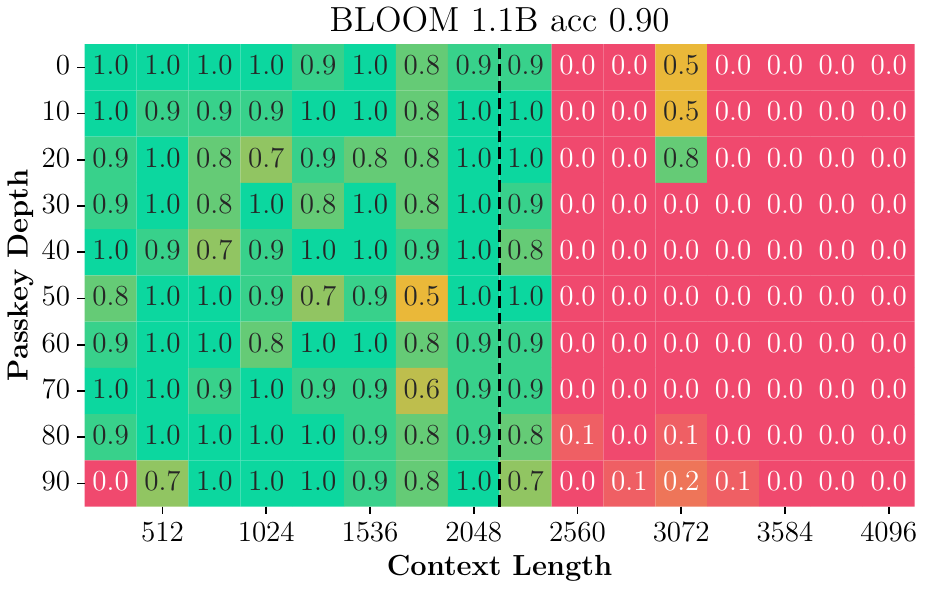}
  \includegraphics[width=0.49\linewidth]{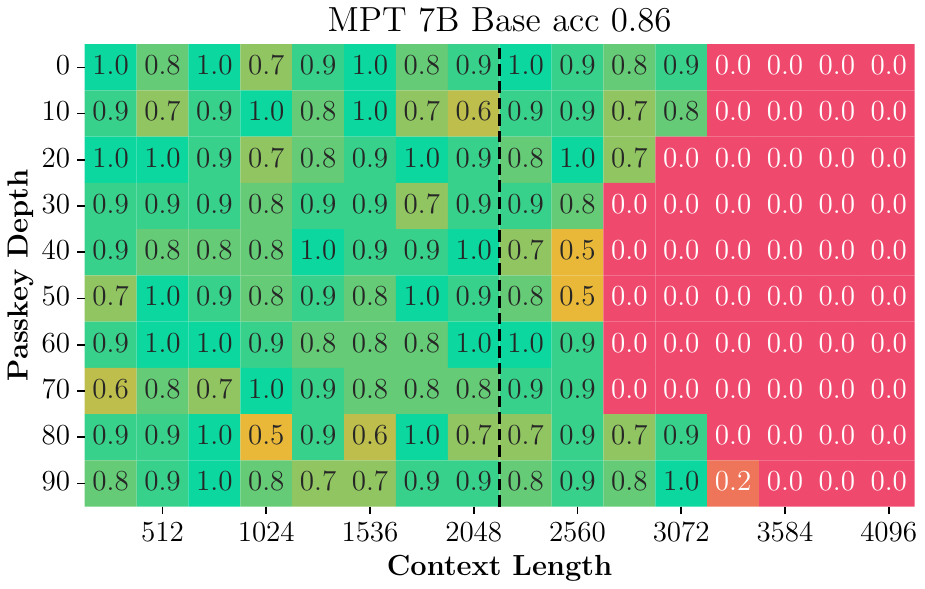}
  \caption{The results of passkey retrieval for ALiBi baselines. The vertical dashed line represents the pre-training length. While ALiBi models do exhibit performance beyond the pre-trained length, their expansion is not substantial. }
  \label{fig:pk_alibi}
\end{figure*}

%% file: fig/fig_ablation_pk.tex
\begin{figure*}[t]
  \centering
  \includegraphics[width=0.49\linewidth]{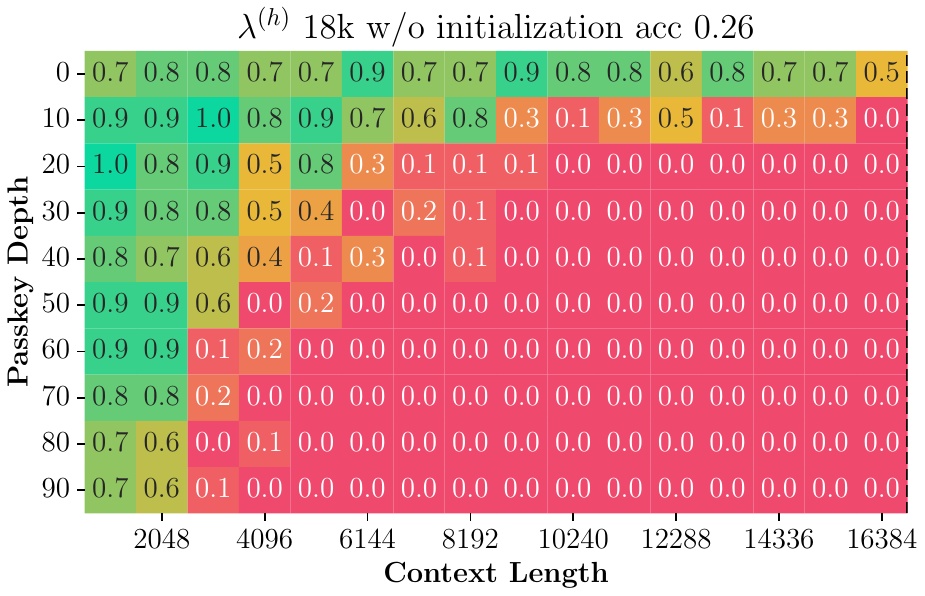}
  \includegraphics[width=0.49\linewidth]{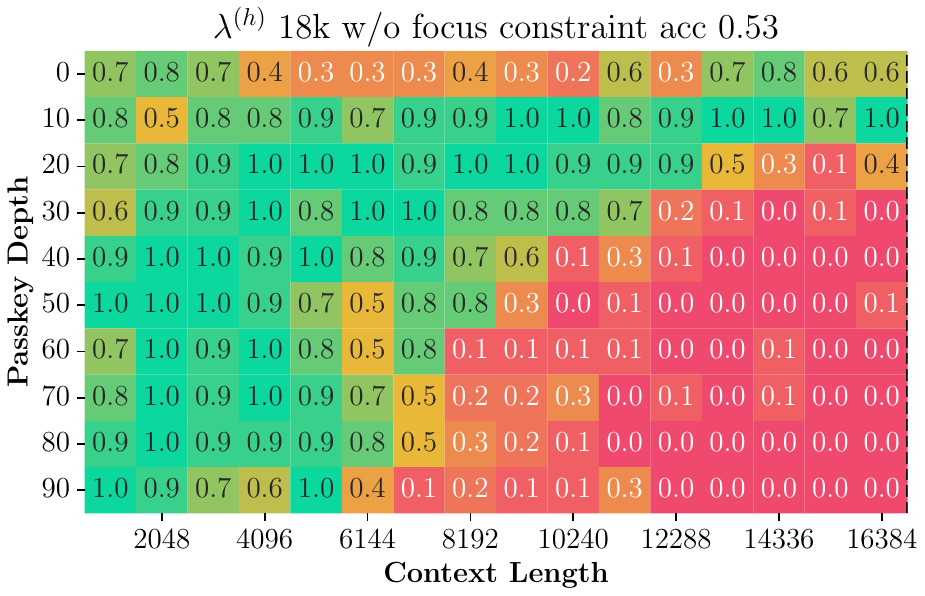}
  \caption{The results of passkey retrieval for HeadScale variations. These results are anticipated to apply to a context length of 16K, but they fail to retrieve the passkey unless it is positioned at the beginning of the context window.}
  \label{fig:ablation_pk}

\end{figure*}

%% file: fig/fig_all_heads_ent_8k.tex
\begin{figure*}[t]
  \centering
  \includegraphics[width=\textwidth]{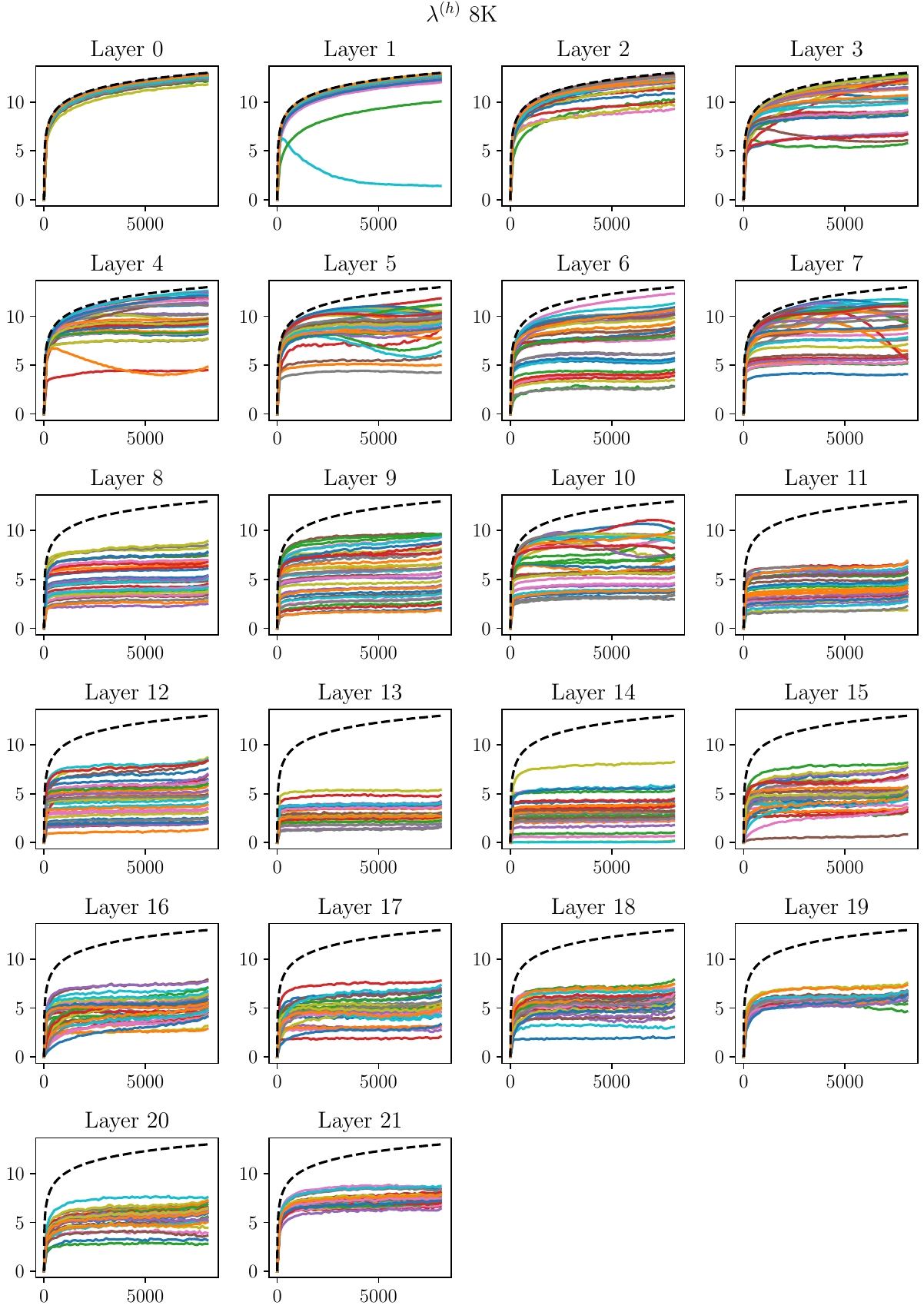}
  \caption{Entropy across all layers and all head of 8k extension head-based scale method. The x-axis is the position of extension and the y-axis is entropy averaged over all test samples. The black dashed curve is the theoretical upper-bound of entropy.}
  \label{fig:all_heads_8k}

\end{figure*}

%% file: fig/fig_all_heads_ent_uni.tex
\begin{figure*}[t]
  \centering
  \includegraphics[width=\textwidth]{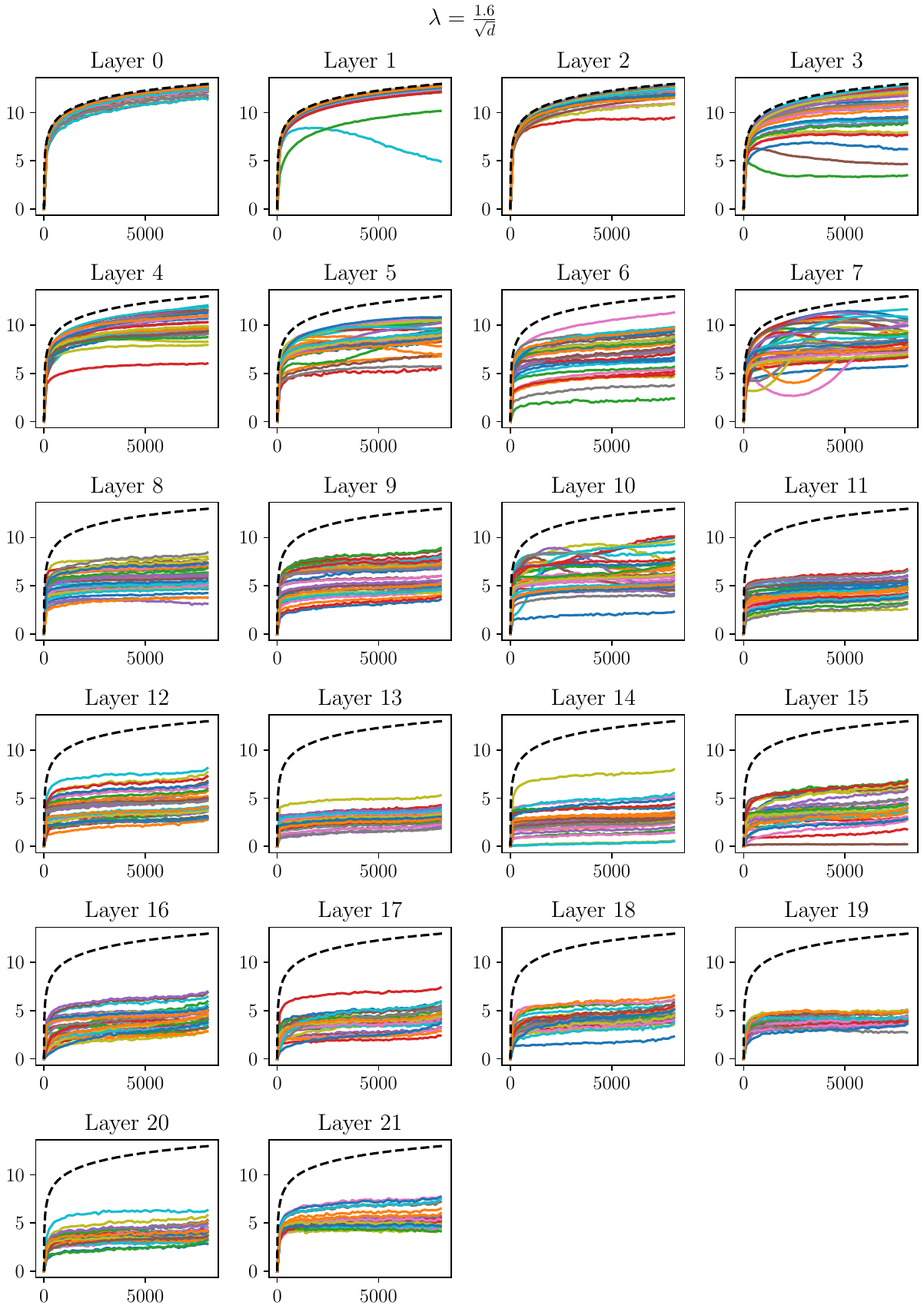}
  \caption{Entropy across all layers and all head of UniformScale with $\lambda=\frac{1.6}{\sqrt d}$}
  \label{fig:all_heads_uni}

\end{figure*}

%% file: fig/fig_all_heads_ent_ori.tex
\begin{figure*}[t]
  \centering
  \includegraphics[width=\textwidth]{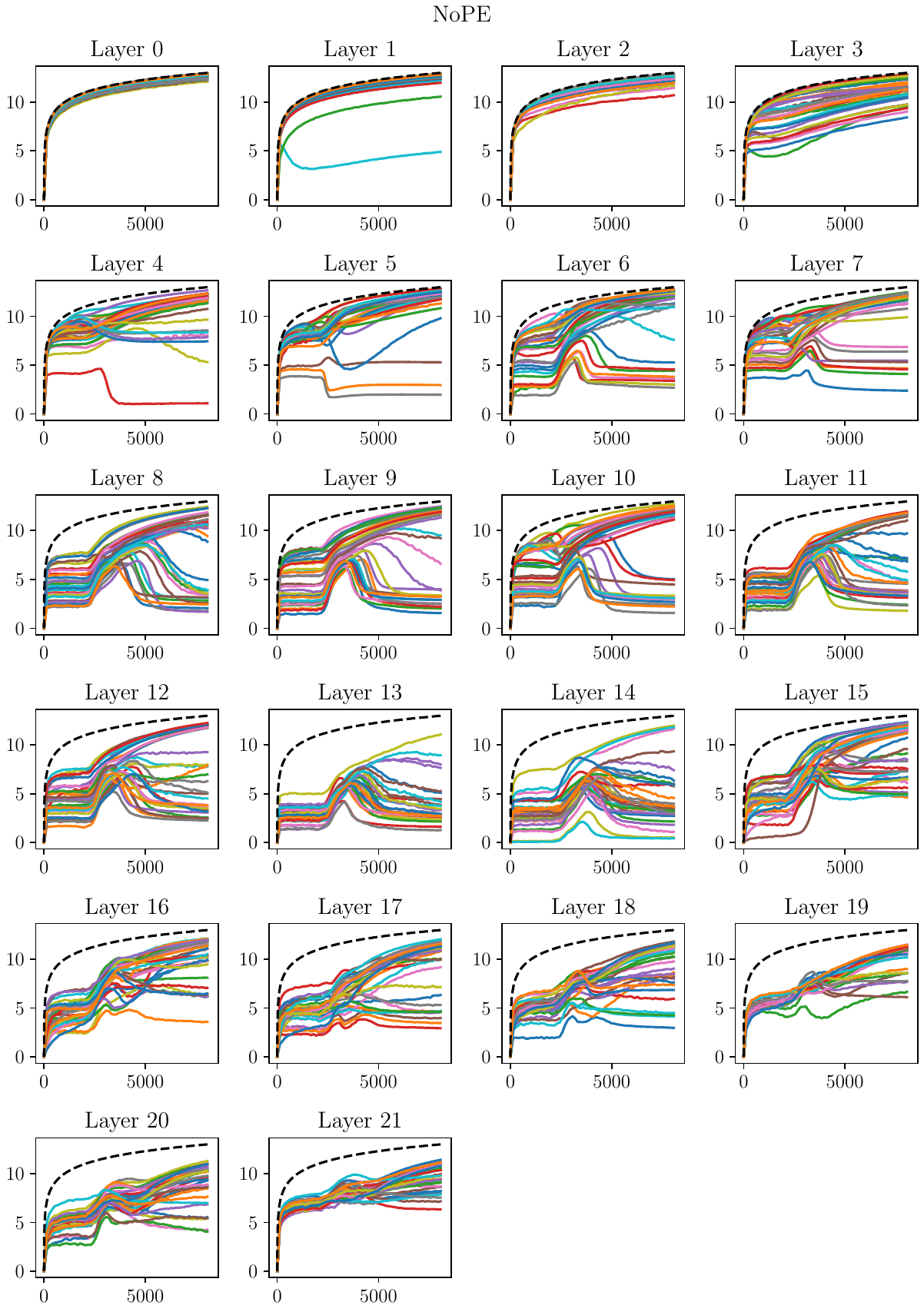}
  \caption{Entropy across all layers and all head of the original NoPE.}
  \label{fig:all_heads_ori}

\end{figure*}